\pdfoutput=1

\documentclass[11pt]{article}

\usepackage{acl}

\def\code#1{\texttt{#1}}
\usepackage{xspace}

\usepackage{array, boldline, makecell, booktabs}
\usepackage[svgnames, table]{xcolor}
\usepackage{multirow}
\newcommand{\thickhline}{\hlineB{4}}

\makeatletter
\NewDocumentCommand{\supptitle}{s}{
\twocolumn[{
  \begin{center}
    \vspace*{-0.3cm}
    \rule{\textwidth}{0.05cm}\\[0.1cm]
    \textbf{- Appendix -}\\[0.2cm]
    {\Large \textbf{\mytitle}}\\[0.1cm]
    \rule{\textwidth}{0.05cm}\\[0.3cm]
  \end{center}
}]
}
\makeatother

\usepackage{tabularx}
\usepackage{capt-of}
\newcommand{\eg}{\emph{e.g.,~}}
\newcommand{\ie}{\emph{i.e.,~}}

\definecolor{LightCyan}{rgb}{0.88,1,1}
\definecolor{Blue}{rgb}{0, 0.5, 1}
\definecolor{Green}{rgb}{0.0, 0.8, 0.0 }
\definecolor{Red}{rgb}{0.95, 0.55, 0.6}
\definecolor{Skyblue}{rgb}{0.6, 0.6, 0.95 }
\definecolor{Beige}{rgb}{0.96, 0.96, 0.86}

\newcommand{\alg}{{NeuRIT}\xspace}
\newcommand{\mytitle}{Where Does Robustness Live? Neuron-Guided Adaptation for Retrieval-Augmented Language Models}

\usepackage[normalem]{ulem}
\useunder{\uline}{\ul}{}

\usepackage{tcolorbox}
\tcbuselibrary{skins, breakable}

\usepackage[hang,flushmargin]{footmisc}

\usepackage{amsmath,amsfonts,bm}

\def\eqref#1{equation~\ref{#1}}

\def\1{\bm{1}}

\DeclareMathAlphabet{\mathsfit}{\encodingdefault}{\sfdefault}{m}{sl}
\SetMathAlphabet{\mathsfit}{bold}{\encodingdefault}{\sfdefault}{bx}{n}

\usepackage{times}
\usepackage{latexsym}
\usepackage{tikz}
\usetikzlibrary{positioning, arrows.meta, shapes.geometric, calc, decorations.pathreplacing}

\usepackage{amsmath}  
\usepackage{amssymb}  
\usepackage[T1]{fontenc}

\usepackage[utf8]{inputenc}

\usepackage{microtype}

\usepackage{inconsolata}

\usepackage{graphicx}

\usepackage{multirow}
\usepackage{booktabs}
\usepackage{graphicx}
\newcommand{\sigdag}{\textsuperscript{\dag}}

\usepackage{algorithm}
\usepackage{footmisc}

\usepackage{newfloat}
\usepackage{listings}
\DeclareCaptionStyle{ruled}{labelfont=normalfont,labelsep=colon,strut=off} 
\floatstyle{ruled}
\newfloat{listing}{tb}{lst}{}
\floatname{listing}{Listing}
\usepackage{times}
\usepackage{latexsym}
\usepackage{mathtools}
\usepackage{amsthm}

\usepackage[capitalize,noabbrev]{cleveref}

\usepackage{tikz}
\usetikzlibrary{fadings}
\usetikzlibrary{patterns}
\usetikzlibrary{shadows.blur}
\usetikzlibrary{shapes}

\usepackage{booktabs}   
\usepackage[table]{xcolor} 
\usepackage{colortbl}  
\definecolor{lightblue}{RGB}{220,235,255}
\usepackage{latexsym}
\usepackage{nccmath}
\usepackage{bbm}
\usepackage{tabularx}
\usepackage{yhmath}
\usepackage{times}
\usepackage{latexsym}
\usepackage{booktabs}
\usepackage{graphicx}
\usepackage{algorithm}
\usepackage{subcaption}
\usepackage{algpseudocode}
\usepackage{amsmath}
\usepackage{siunitx}
\usepackage{mathtools}
\usepackage{array}
\usepackage{multirow}
\usepackage[normalem]{ulem}
\usepackage{arydshln}

\useunder{\uline}{\ul}{}
 
\makeatletter
\newcommand{\multiline}[1]{%
  \begin{tabularx}{\dimexpr\linewidth-\ALG@thistlm}[t]{@{}X@{}}
    #1
  \end{tabularx}
}
\makeatother

\algblock{Input}{EndInput}
\algnotext{EndInput}
\algblock{Output}{EndOutput}
\algnotext{EndOutput}

\title{\mytitle}

\author{
  \textbf{Jae O Lee\textsuperscript{1}}\thanks{Equal contribution.},
  \textbf{Jaemin Kim\textsuperscript{1}}\footnotemark[1],
  \textbf{Sumyeong Ahn\textsuperscript{2}}\thanks{Co-corresponding authors.},
  \textbf{Seo Yeon Park\textsuperscript{1}}\footnotemark[2]
\\
\\
  \textsuperscript{1}Hanyang University,
  \textsuperscript{2}KENTECH
\\
  \texttt{\{drizzly159, jaemkim01, seoyeonpark\}@hanyang.ac.kr}
\\
  \texttt{sumyeongahn@kentech.ac.kr}
}

\begin{document}

\maketitle

\begin{abstract}
Retrieval-Augmented Language Models (RALMs) have shown strong potential in knowledge-intensive tasks, yet they remain vulnerable when retrieved contexts are noisy or irrelevant. Robustness against such contexts requires two distinct capabilities: \textit{abstention} when contexts are uninformative, and \textit{selective extraction} when relevant evidence is buried in noise. Yet existing methods face two key limitations: they do not train separately for these two capabilities, and they adapt the model at a coarse layer- or module-level granularity, overlooking that only a small subset of neurons is strongly activated for a given input. We propose \alg, a \textbf{Neu}ron-guided \textbf{R}obust \textbf{I}nstruction-\textbf{T}uning framework built on a \textit{localization-first} perspective. \alg mines context-aware neurons associated with relevant and irrelevant context processing, and uses them as \textit{anchors} to selectively adapt both the identified neuron groups and the layers in which they concentrate. \alg then performs two-stage instruction tuning that teaches complementary behaviors: suppress generation when there is nothing to extract, and extract relevant evidence when there is. \alg consistently outperforms strong baselines across diverse QA benchmarks and generator backbones. Our code is available at \url{https://github.com/HYU-ARK-Lab/NeuRIT}.

\end{abstract}

\section{Introduction}
\label{sec:intro}

While Large Language Models (LLMs) have achieved remarkable success, they still face significant limitations, including their tendency to produce hallucinations in knowledge-intensive tasks (\eg question-answering)~\cite{wang-yu-2025-iquest, huang2025survey}. To mitigate these problems, Retrieval-Augmented Language Models (RALMs)~\cite{lewis2020retrieval} provide a promising solution. RALMs operate in two main steps: (1) invoking a retriever to search for and extract relevant contexts to a given query from external sources, and (2) leveraging the retrieved contexts to prompt LLMs, thereby generating more accurate and grounded responses~\cite{guu2020retrieval, izacard2023atlas, gao2023retrieval, fan2024survey}. Despite their ability to support grounded generation, RALMs face a significant challenge: query-irrelevant or incorrectly retrieved contexts can mislead the LLM and consequently degrade its performance~\cite{wang-etal-2023-self-knowledge, cuconasu2024power, chen2024benchmarking}, as LLMs are not explicitly trained to be robust against such noise~\cite{jiang2024llms, lin2025llm}.

Importantly, retrieved contexts can fail in two qualitatively distinct ways, each demanding a fundamentally different model behavior. They may be {uninformative} to the query, providing no usable evidence at all, in which case the model must abstain from extracting anything and rely on a refusal-like response. Alternatively, they may be {partially relevant}, containing the answer alongside distracting noise, in which case the model must selectively extract the useful fragment while ignoring the surrounding noise. A model trained without this distinction risks either over-suppressing useful evidence or under-suppressing noise. Robust generation thus demands two distinct capabilities: \textit{abstention} and \textit{selective extraction}.

Existing approaches have attempted to address this challenge in various ways. 
While prior work has explored refining retrieved contexts before generation~\cite{glass-etal-2022-re2g, vig-etal-2022-exploring, xu2023recomp, wu2025rankcot}, a more direct line of work enhances the robustness of the generator LLM itself through instruction tuning~\cite{wei2024instructrag, lin2023ra, wu-etal-2025-pa}, teaching the model to better utilize retrieved contexts while disregarding query-irrelevant ones. 
However, we observe two key limitations. \textit{First, in terms of the training objective}, existing methods do not train separately for \textit{abstention} and \textit{selective extraction}, leaving the tension between the two unresolved.  \textit{Second, in terms of the granularity of adaptation}, they typically update parameters at the layer, module, or low-rank level~\cite{wu-etal-2025-pa, yoran2024makingretrievalaugmentedlanguagemodels, xu2024parenting}, neglecting a well-established property of LLMs: only a small subset of neurons is strongly activated for a given task or input~\cite{li2022lazy, xu2025let}. This suggests that robustness-related computations might be concentrated in limited regions of the network, and updating parameters without this consideration may unnecessarily interfere with unrelated capabilities. 
One possible solution is to identify robustness-related neurons and update them selectively. However, neuron-level updates alone are unlikely to reliably reshape model behavior, since model computation is inherently distributed across many parameters~\cite{leng2025towards}. We therefore use identified neurons not as isolated update targets, but as \textit{anchors} for adaptation: neuron-level attribution identifies context-aware behaviors underlying robustness (\ie processing relevant or irrelevant contexts), and guides updates to both these neuron groups and the layers in which they are densely concentrated.

To this end, we propose \alg (\underline{\textbf{Neu}}ron-guided \underline{\textbf{R}}obust \underline{\textbf{I}}nstruction \underline{\textbf{T}}uning), an instruction-tuning framework built on a localization-first perspective that addresses both limitations above. To address the granularity limitation, \alg first performs \textit{context-aware neuron mining}, which identifies neurons (\ie existing in FFN) associated with relevant and irrelevant context processing through attribution-based analysis. 
Since some neurons respond to both context types, we separate them into a shared group rather than assigning them to either context-specific group, preserving the discriminative signals of relevant- and irrelevant context-aware neurons. 
To separate these objectives, \alg then performs \textit{neuron-guided instruction tuning}, which teaches the model two complementary behaviors---suppression and evidence extraction---through two training stages. In the suppression stage, \alg updates only irrelevant context-aware neurons to suppress generation under uninformative contexts. In the extraction stage, \alg further adapts all three identified neuron groups and the layers in which they concentrate, teaching the model to extract relevant evidence from partially noisy contexts. Together, these stages yield a unified robustness capability: suppress when there is nothing to extract, and extract when there is.


Our contributions are summarized as follows:
\begin{itemize}
    \item We introduce a new perspective on robustness-oriented adaptation for RALMs: neuron attribution provides a localization signal for identifying where adaptation should occur.

    \item We propose \alg, a two-stage neuron-guided instruction-tuning framework that uses mined context-aware neurons as \textit{anchors} to jointly adapt the identified neuron groups and the layers in which they concentrate, while training separately for two complementary behaviors: suppression under uninformative contexts and selective extraction from partially noisy contexts.

    \item Extensive experiments on various QA benchmarks and different generator backbones show consistent gains over strong baselines under both in-domain and out-of-domain settings. Our analysis further reveals that neuron-guided localization drives the gains.
\end{itemize}

\section{Related Work}
\paragraph{Refining Retrieval Contexts.}
Prior works have proposed various context refinement methods to enhance the quality of retrieved knowledge. For example, reranking retrieved documents~\cite{glass-etal-2022-re2g}, extracting query-related content from the retrieved documents (\ie summarization) ~\cite{vig-etal-2022-exploring, xu2023recomp}, and iterative retrieval along with intermediate reasoning paths~\cite{trivedi2023interleaving, wang2024rat, wu2025rankcot} have been explored. Recently, ~\citet{wu2025rankcot} proposed an advanced knowledge refinement method that incorporates re-ranking signals in generating Chain-of-Thought (CoT)-based summarization. However, many existing works have created extra modules for the knowledge refinement, which introduce additional steps beyond generation on RALM. These additional steps introduce extra computational costs, making it less practical for large-scale deployment~\cite{wu-etal-2025-pa}. Moreover, these additional steps may suffer from cascading errors~\cite{yu-etal-2022-retrieval}, which ultimately limit their effectiveness in real-world applications.

\paragraph{Robustness to Retrieved Contexts in RALM.}
Previous studies have sought to improve the robustness of RALMs to noisy or irrelevant retrieved contexts. \citet{yoran2024makingretrievalaugmentedlanguagemodels} train language models to directly generate correct answers despite input noise, implicitly enhancing robustness. More recent works leverage instruction tuning to make the denoising process explicit: \citet{wei2024instructrag} prompts the model to synthesize rationales that articulate the reasoning path from noisy contexts to the ground-truth answer, while \citet{wu-etal-2025-pa} performs instruction tuning on data refined through a citation rewrite mechanism that uses NLI models to ensure the model cites only valid evidence. However, these approaches share two limitations. First, they collapse the qualitatively distinct failure modes of retrieval, fully irrelevant versus partially relevant contexts, into a single training objective, leaving the tension between abstention and selective extraction unresolved. Second, they rely on dense optimization strategies such as full-parameter fine-tuning or layer-level parameter-efficient fine-tuning, overlooking the neuron-level sparsity of LLMs and thus failing to selectively update the specific neurons critical for noise robustness.

\section{Methodology}
\label{method}
\begin{figure*}[t]
    \centering
    \includegraphics[width=1.0\textwidth, trim={0 0.3cm 0 0.5cm}, clip]{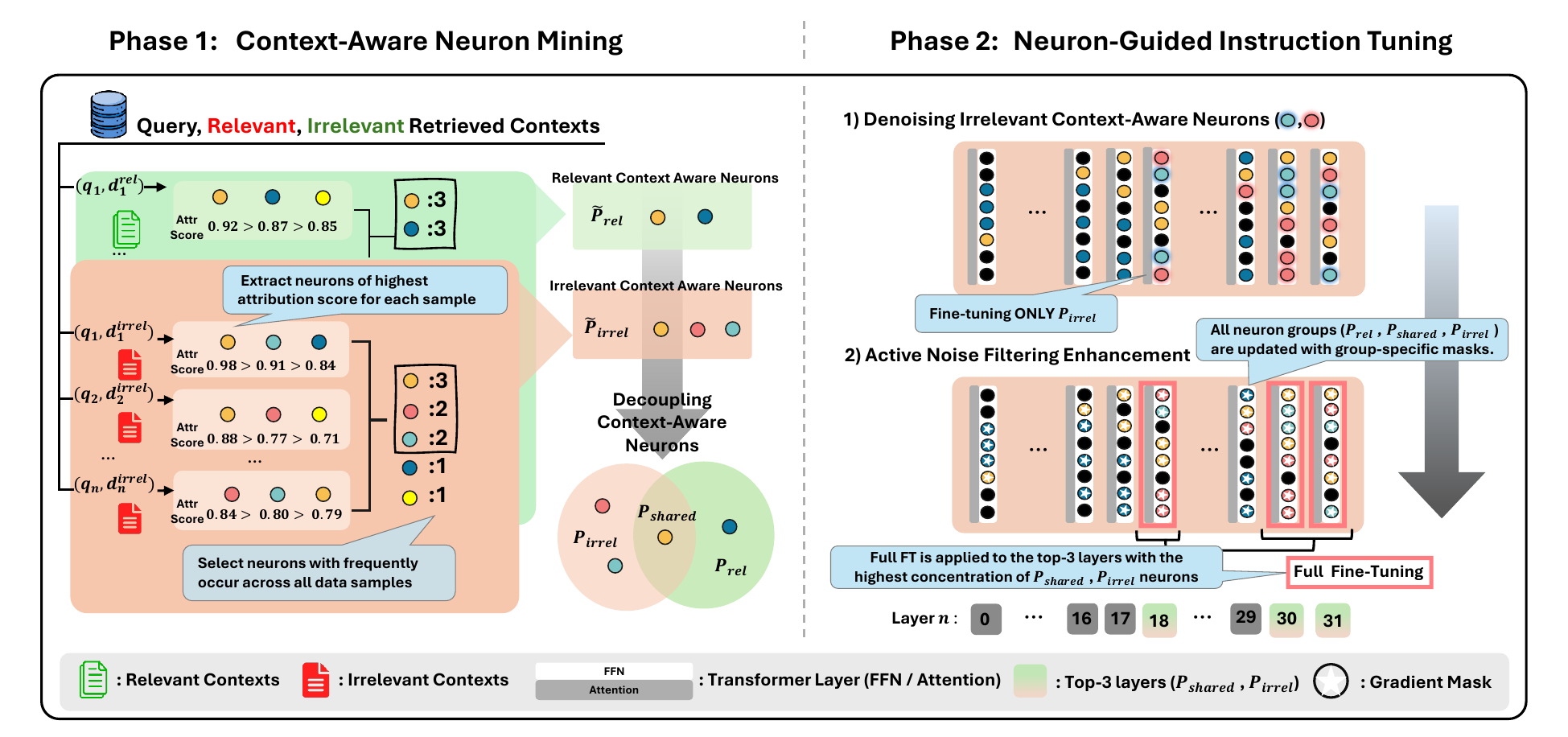}
    \vspace{-4mm}
    \caption{Overview of the \alg framework. \alg consists of two phases: (1) Context-Aware Neuron Mining, where neurons are categorized into relevant, irrelevant, and shared groups based on attribution scores from retrieved contexts; and (2) Neuron-Guided Instruction Tuning, where the identified neuron groups guide the training process through group-specific updates.}
    \label{fig:mining}
\end{figure*}

\paragraph{Problem Setup.} 
Retrieval-Augmented Language Models (RALMs) for knowledge-intensive tasks, such as question-answering, aim to generate answers using a generator LLM, based on a query and documents returned by a retriever. 
Formally, let $\mathcal{D}_{\text{Q}} = \left\{(q_i, \{d_{i,j}\}_{j=1}^{k})\right\}_{i=1}^{N}$ be a set of open-domain questions, where $q_i$ denotes a query and $\{d_{ij}\}_{j=1}^k$ is the set of top-$k$ retrieved documents associated with $q_i$.
Note that the documents are ranked based on relevance score with the given query. 
Then, a generator LLM $\mathcal{M}$ is expected to produce the output $\hat{y}$ as follows: $\hat{y}_i = \mathcal{M}(\code{Prompt}_i)$ where $\code{Prompt}_i=(q,\{d_{i,j}\}_{j=1}^k)$ refers to a predefined prompt template along with inputs. 
However, the retrieved documents set $\{d_{i,j}\}_{j=1}^{k}$ often contain query-irrelevant documents that lead to performance degradation. To this end, we propose to identify relevant and irrelevant context-aware neurons and enhance their robustness to retrieved information. Our method consists of two primary components: 1) context-aware neuron mining, and 2) robustness enhancement instruction tuning on the identified neurons. The overall architecture is illustrated in Figure~\ref{fig:mining}.

\subsection{Context-Aware Neuron Mining} 
\paragraph{Attribution-Based Context-Aware Neuron Identification.}
To decide where robustness-oriented adaptation should occur, we first quantify which FFN neurons contribute to processing relevant and irrelevant retrieved contexts. 
To quantify the contribution of each neuron to processing relevant and irrelevant contexts, we leverage the attribution method of \citet{shi2024ircan}.
For each query $q_i$, we first construct a triplet $(q_i, d_i^{\textrm{rel}}, d_i^{\textrm{irrel}})$. Here, $d_i^{\textrm{rel}}$ denotes a top-ranked relevant document, and $d_i^{\textrm{irrel}}$ denotes an irrelevant document with a zero relevance score. Next, for every query $q_i$ and document $d_i^c$ with context type $c \in \{\text{rel,irrel}\}$, we measure the attribution score of every neuron $n$ (that exists in FFNs) to the prediction $y$. This is done via Integrated Gradients~\cite{sundararajan2017axiomatic}.
Specifically, let $v(q_i)$ denote the activation of neuron $n$ given only the query, and $v(q_i, d_i^c)$ denote the activation given both the query and the document of context type $c$. We compute the attribution score as follows: 
\begin{equation}
\begin{split}
    \mathcal{A}(n; q_i, d_i^c)
    = 
    \bigl(v(q_i,d_i^c)-v(q_i)\bigr) \\ \times
    \int_{\alpha=0}^1
    \frac{\partial P(y \mid q_i, d_i^c, v_{\alpha})}
    {\partial v_{\alpha}}
    \, d\alpha ,
\end{split}
\label{eq:attr}
\end{equation}
where $v_{\alpha} = v(q_i) + \alpha \bigl(v(q_i,d_i^c)-v(q_i)\bigr)$ interpolates between the two activation states for $\alpha \in [0,1]$, and the integral is approximated using a 20-step Riemann sum.
Based on these scores, for each query, we filter for the top 10\% and select the top 20 highest-scoring neurons. 
We then aggregate these neuron selections across all queries within each context type, and identify the top-$t$ most frequently occurring neurons. We denote these initial candidate sets of neurons as $\tilde{\mathcal{P}}_{\text{irrel}}$ and $\tilde{\mathcal{P}}_{\text{rel}}$ for irrelevant and relevant contexts, respectively (See Figure~\ref{fig:mining}, left). 

\paragraph{Decoupling Context-Aware Neurons.}
Once we obtain $\tilde{\mathcal{P}}_{\text{rel}}$ and  $\tilde{\mathcal{P}}_{\text{irrel}}$, we explore the intersection between the two sets. 
We observe that identified neuron sets $\tilde{\mathcal{P}}_{\text{irrel}}$ and $\tilde{\mathcal{P}}_{\text{rel}}$ reveal a significant overlap\footnote{Initial mining yields 7,380 relevant and 7,036 irrelevant context-aware candidate neurons with a significant overlap of 6,240 neurons.}. 
This is natural since parameters associated to neurons in LLMs form distributed and overlapping representations rather than being functionally exclusive \cite{leng2025towards}. 
However, such overlapping neurons do not provide discriminative signals for identifying context-specific awareness, as their activations cannot be uniquely attributed to either relevant or irrelevant contexts. 
To address this, we define distinct neuron sets by excluding neurons that exhibit high attribution in both context types. 
Specifically, we define the overlapping neurons set as $\mathcal{P}_{\text{shared}}=\tilde{\mathcal{P}}_{\text{rel}} \cap \tilde{\mathcal{P}}_{\text{irrel}}$, and remove them from each set as follows:
$\mathcal{P}_{\text{rel}} \leftarrow \tilde{\mathcal{P}}_{\text{rel}} \setminus \mathcal{P}_{\text{shared}}$,
$\mathcal{P}_{\text{irrel}} \leftarrow \tilde{\mathcal{P}}_{\text{irrel}} \setminus \mathcal{P}_{\text{shared}}$.
This procedure ensures that the resulting neuron sets capture distinct context-specific awareness for relevant and irrelevant contexts, respectively (See \autoref{fig:mining}, left). Our mining results in 100 neurons for $\mathcal{P}_{\text{rel}}$ and $\mathcal{P}_{\text{irrel}}$ and 30 neurons for $\mathcal{P}_{\text{shared}}$.

\subsection{Neuron-Guided Instruction Tuning} 
\label{prompt_tuning}
Based on the identified neurons, we propose a two-stage neuron-guided instruction tuning framework, with each stage targeting one robustness behavior: (1) a denoising stage that teaches abstention by suppressing irrelevant context-aware neurons under uninformative contexts, and (2) a noise filtering stage that teaches selective extraction by adapting the identified neuron groups within the layers where noise-related neurons are most densely concentrated.

\paragraph{Denoising Irrelevant Context-Aware Neurons.}
For uninformative retrieved contexts, robust generation requires abstention rather than producing answers from unsupported evidence.
To induce this behavior at the neuron level, we instruction-tune the irrelevant context-aware neurons $\mathcal{P}_{\text{irrel}}$ to emit an End-of-Text (\texttt{EOT}) token under irrelevant-context inputs.
Since $\mathcal{P}_{\text{irrel}}$ are identified as primarily responsible for processing irrelevant or misleading contexts, allowing them to participate in processing relevant information poses a risk of latent interference. 
Accordingly, we impose a hard constraint that decouples these irrelevant context-aware neurons from generation.

\paragraph{Noise Filtering Enhancement on Neuron-level Parameters.}
In partially relevant retrieval settings, abstention alone is insufficient because useful evidence may be mixed with distracting noise. 
We therefore adapt the regions guided by the identified neurons, targeting where noise-robustness computations are concentrated. 
This is because neuron-level updates alone are unlikely to reliably reshape model behavior, since model computation is inherently distributed across many parameters~\cite{leng2025towards}. 
Hence, we update not only the identified neuron groups ($\mathcal{P}_{\text{rel}}$, $\mathcal{P}_{\text{shared}}$, $\mathcal{P}_{\text{irrel}}$), but also the top-3 layers with the highest density of $\mathcal{P}_{\text{irrel}}$ and $\mathcal{P}_{\text{shared}}$, as these layers aggregate noise-discriminative signals near the output\footnote{See $\S$~\ref{para:layerdistribution} for the layer-wise distribution analysis motivating this choice.}. 
This allows the model to refine relevant-evidence extraction while controlling interference from irrelevant contexts. 
Specifically, we optimize the selected parameters to make the generator produce a distilled summary that contains only query-relevant information from potentially noisy retrieved contexts. Given an original query and its retrieved contexts, the model is instructed to filter out irrelevant content and generate only the relevant evidence needed to answer the query. This aligns the localized adaptation with our context-filtering objective. For this, we construct a dataset as follows: 
for a pair $(q_i, \{d_{i,j}\}_{j=1}^k) \in \mathcal{D}_{Q}$, we employ an external LLM\footnote{In our experiments, we used \code{GPT-4.1.mini} (OpenAI 2025)\footnote{~https://openai.com/index/gpt-4-1/} though any LLM can be used.} to generate Relevant Summaries (RS) $y_{i}^{\text{RS}}$, which includes only query-relevant information from $\{d_{ij}\}_{j=1}^k$ by giving a specific instruction\footnote{\label{fn:shared}~See Table~\ref{data_construct} in Appendix for details}.   
We construct an instruction-tuning dataset as follows:
\begin{equation*}
\mathcal{D}_{\text{RS}} = \left\{\left(q_i, \{d{_{ij}}\}_{j=1}^{k}),\ y_i^{\text{RS}}\right)\right\}_{i=1}^{n}
\end{equation*}
Once done training, to enable open-domain question answering, we design a dual-instruction prompt consisting of (1) relevant summary extraction (\code{inst}$_{\text{RSE}}$: \texttt{extract relevant information from provided documents}) and (2) a question answering instruction ($\code{inst}_{\text{QA}}$: \texttt{Answer a question as briefly as possible}). Each training input $x_i$ is formatted as a prompt combining these components:
\vspace{-2mm}

\begin{equation*}
\vspace{-2mm}
\code{Prompt}_i = (\code{inst}_{\text{RSE}}, \code{inst}_{\text{QA}}, \{d_{ij}\}_{j=1}^{k}, q_i)
\end{equation*}

\noindent The instruction-tuned generator LLM produces the corresponding answer as $\hat{y}_i = \mathcal{M}(\code{Prompt}_i)$. 
This design jointly enables the model to filter noisy documents, extract and align relevant evidence, and generate factually grounded answers.
\vspace{-2mm}

\section{Experiments}
\label{sec:experimental_settings}

\subsection{Data}
\label{sec:data}

We evaluate our method on a diverse QA benchmarks covering various question types and reasoning settings. We use KILT-formatted versions of Natural Questions, TriviaQA, and HotpotQA \cite{rau2024bergen}.
\textbf{KILT-NQ (Natural Questions)}~\cite{kwiatkowski2019natural} contains real anonymized Google search queries with answer annotations on Wikipedia pages.
\textbf{ASQA}~\cite{stelmakh2022asqa} is a long-form QA benchmark for ambiguous factoid questions, requiring answers to synthesize evidence from multiple sources and resolve different interpretations.
\textbf{KILT-TriviaQA}~\cite{joshi2017triviaqa} is a factoid QA benchmark derived from TriviaQA, with answer provenance aligned to Wikipedia evidence.
\textbf{SciQ}~\cite{welbl2017crowdsourcing} is a multiple-choice science QA dataset consisting of crowdsourced exam-style questions with supporting evidence paragraphs.
\textbf{POPQA}~\cite{mallen2023not} evaluates factual knowledge about entities with varying popularity, especially long-tail entities.
\textbf{KILT-HotpotQA}~\cite{yang2018hotpotqa} requires multi-hop reasoning over multiple Wikipedia pages.
\textbf{2WikiMultiHopQA}~\cite{ho2020constructing} requires connecting information across multiple Wikipedia articles to answer entity-centric multi-hop questions.

\subsection{Baselines}
\label{sec:baselines}

To validate the effectiveness of \alg, we conduct a comparative evaluation against three distinct categories of baselines: 1) Standard RALM, 2) {Robustness Enhancing Tuning for RALMs}, and 3) {Retrieved Context Refinement}.
For the refinement methods, we additionally evaluate them in a compositional manner by attaching each refinement module to every baseline in the first two categories.

\paragraph{Standard RALM.} This baseline follows the conventional Retrieval Augmented Language Model (RALM) generation paradigm, where retrieved documents are concatenated along with a given query without any additional control mechanisms. 

\paragraph{Robustness Enhancing Tuning for RALMs.}
These methods tune the {generator} LLM directly to be robust to irrelevant or misleading retrieved contexts: \textbf{RetRobust}\footnote{https://huggingface.co/Ori/llama-2-13b-peft-nq-retrobust} \cite{yoran2024makingretrievalaugmentedlanguagemodels} trains the generator LLM on noisy retrieval inputs to directly produce correct answers despite distracting or misleading contexts, encouraging robustness to irrelevant evidence without explicit denoising supervision.
\textbf{InstructRAG}\footnote{https://huggingface.co/meng-lab/2WikiMultiHopQA-InstructRAG-FT} \cite{wei2024instructrag} instead applies instruction tuning on self-synthesized rationales that explain how ground-truth answers are derived from noisy retrieved documents, providing explicit denoising supervision. 
\textbf{PA-RAG}\footnote{https://huggingface.co/wuqiong1/PA-RAG\_Meta-Llama-3-8B-Instruct} \cite{wu-etal-2025-pa} combines instruction fine-tuning with multi-stage DPO that sequentially optimizes response informativeness, response robustness, and citation quality.

\begin{table*}[t]
\centering
\scriptsize
\renewcommand{\arraystretch}{1.3}
\setlength{\tabcolsep}{5.0pt}

\begin{tabular}{lcccccccc}
\toprule
\textbf{Method} &
\textbf{NQ} &
\textbf{ASQA} &
\textbf{TriviaQA} &
\textbf{SCIQ} &
\textbf{POPQA} &
\textbf{HotpotQA} &
\textbf{2Wiki} &
\textbf{Avg.} \\
\midrule
RALM \cite{lewis2020retrieval} 
& 62.21 & 67.82 & 88.60 & 53.30 & 60.85 & 45.44 & 42.27 & 60.07 \\
\cdashline{1-9}
\quad + Reranker \cite{glass-etal-2022-re2g}
& 61.65 & 67.93 & 88.31 & 52.60 & 60.37 & 45.27 & 41.84 & 59.71 \\
\quad + RankCoT \cite{wu2025rankcot}
& 59.43 & 65.93 & 89.60 & 57.40 & 59.80 & 48.54 & 43.42 & 60.59 \\
\midrule
RetRobust \cite{yoran2024makingretrievalaugmentedlanguagemodels}
& 53.23 & 61.92 & 89.80 & 58.20 & 53.24 & 40.84 & 35.17 & 56.06 \\
\cdashline{1-9}
\quad + Reranker \cite{glass-etal-2022-re2g}
& 54.42 & 64.13 & 90.55 & 58.20 & 54.92 & 41.46 & 35.69 & 57.05 \\
\quad + RankCoT \cite{wu2025rankcot}
& 53.26 & 60.55 & 88.93 & 57.80 & 53.18 & 43.84 & 39.94 & 56.79 \\
\midrule
InstructRAG \cite{wei2024instructrag}
& 60.62 & 66.98 & 87.48 & 55.30 & 60.25 & 45.33 & 44.61 & 60.08 \\
\cdashline{1-9}
\quad + Reranker \cite{glass-etal-2022-re2g}
& 63.58 & 71.51 & 90.73 & 56.80 & 64.60 & 48.32 & 47.01 & 63.22 \\
\quad + RankCoT \cite{wu2025rankcot}
& 60.17 & 67.72 & 90.76 & 59.50 & 62.24 & 51.66 & 50.06 & 63.16 \\
\midrule
PA-RAG \cite{wu-etal-2025-pa}
& 67.92 & 74.47 & 89.86 & 56.60 & 64.37 & 49.29 & 46.57 & 64.15 \\
\cdashline{1-9}
\quad + Reranker \cite{glass-etal-2022-re2g}
& \underline{72.01} & \underline{79.43} & \underline{92.58} & 57.60 & \underline{71.08} & \underline{54.12} & 48.87 & \underline{67.96} \\
\quad + RankCoT \cite{wu2025rankcot}
& 66.30 & 73.10 & 92.32 & \textbf{63.40} & 65.68 & 53.89 & \textbf{52.02} & 66.67 \\
\midrule
\midrule
\rowcolor{lightblue} \textbf{\alg}
& 69.68 & 75.31 & 92.21 & 60.80 & 67.37& 52.25 & 46.34 & 66.28\\

\rowcolor{lightblue} \quad + Reranker \cite{glass-etal-2022-re2g}
& \textbf{72.57} & \textbf{80.48} & \textbf{93.91}\sigdag & \underline{61.60} & \textbf{74.29}\sigdag & \textbf{55.71}\sigdag & 47.80 & \textbf{69.48} \\
\rowcolor{lightblue} \quad + RankCoT \cite{wu2025rankcot}
& 61.93 & 69.09 & 91.60 & 59.10 & 63.28 & 52.04 & \underline{51.13} & 64.02 \\
\bottomrule
\end{tabular}
\vspace{-2mm}
\caption{Main results of performance comparison across QA benchmarks.
We report \textbf{Accuracy} for each dataset.
$^{\dagger}$: \alg significantly improves the best baseline at $p < 0.05$ with paired t-test.
The best results are highlighted in \textbf{bold} and the second-best results are \underline{underlined} (per metric).}
\label{tab:main_results}
\end{table*}

\paragraph{Retrieved Contexts Refinement.}
In contrast to methods that directly tune the generator LLM for robustness, these baselines improve robustness by refining retrieved contexts before they are given to the generator LLM. 
\textbf{Reranker}\footnote{https://huggingface.co/naver/trecdl22-crossencoder-debertav3}~\cite{glass-etal-2022-re2g} uses a cross-encoder re-ranker to prioritize query-relevant documents, thereby reducing the influence of irrelevant contexts in the prompt. 
\textbf{RankCoT}\footnote{https://huggingface.co/MignonMiyoung/RankCoT} \cite{wu2025rankcot} instead refines retrieved documents by combining re-ranking signals with CoT-based summarization, serving as a stronger noise-mitigation baseline at the cost of additional reasoning overhead.

\subsection{Implementation Details}
\label{sec:exp_setup}
All experiments are conducted using the generator LLM as {LLaMA-3-8B-Instruct}~\cite{dubey2024llama}, and training is performed on a single NVIDIA H200 GPU with 141GB memory. For all results, we utilize greedy decoding by setting the temperature to 0 to ensure reproducibility. 
To identify context-aware neurons, we use approximately 400 sub-samples of {HotpotQA} training data. We set frequency occurrence threshold $k$ as 130 to ensure 100 neurons are selected for both $\mathcal{P}_{\text{rel}}$ and $\mathcal{P}_{\text{irrel}}$ and 30 neurons for $\mathcal{P}_{\text{shared}}$.
For $\mathcal{D}_{\text{RS}}$, the external LLM produces 142-token summaries grounded exclusively in the provided documents, with no intrinsic knowledge use enforced via prompting to mitigate hallucination. 
Detailed training procedures, prompts, implementation details, and hyperparameter settings for neuron selection are provided in Appendix~\ref{appendix:training_details}. 

\begin{figure*}[t]
    \centering
    \includegraphics[width=0.49\textwidth]{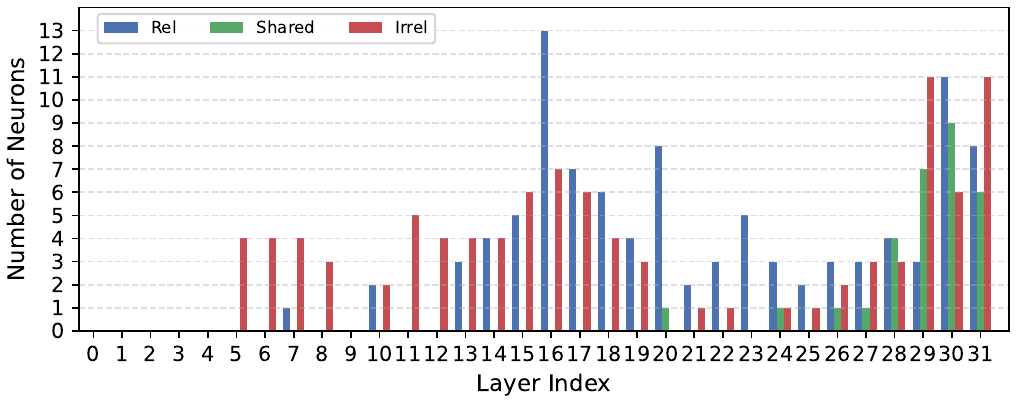}
    \includegraphics[width=0.49\textwidth]{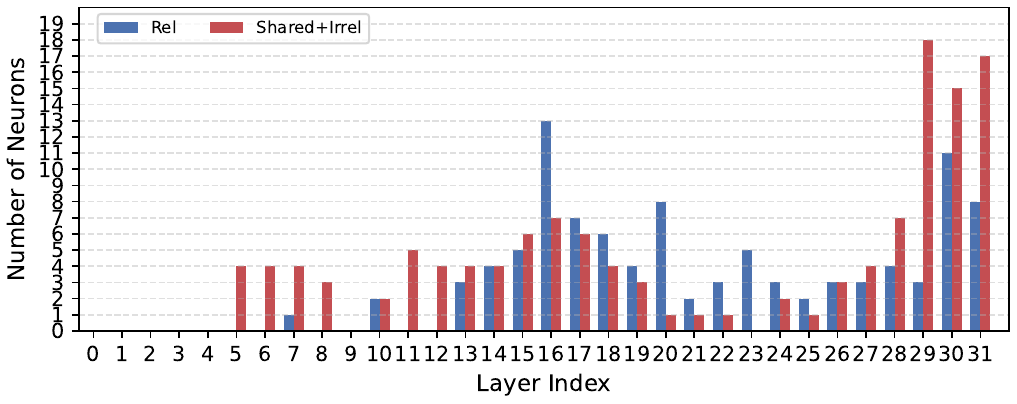}
    \vspace{-4mm}
    \caption{Neuron distribution of LLaMA-3-8B-Instruct on the {HotpotQA} dataset; the left panel shows the distribution of $\mathcal{P}_{\text{rel}}$, $\mathcal{P}_{\text{shared}}$ and $\mathcal{P}_{\text{irrel}}$, while the right panel shows the $\mathcal{P}_{\text{rel}}$ and the combination of $\mathcal{P}_{\text{irrel}}$ and $\mathcal{P}_{\text{shared}}$.}
    \label{fig:neuron_where}
\end{figure*}

\subsection{Main Results}
\label{sec:Result}
We report results in Table~\ref{tab:main_results} for three variants: the base model, the base model with {Reranker}, and the base model with {RankCoT}.
Without ranking, \alg consistently outperforms most baselines and remains comparable to or slightly better than the strongest one, {PA-RAG}, across all datasets.
Notably, this strong base performance is observed despite \alg being trained exclusively on {HotpotQA}, indicating that the learned behavior generalizes beyond the training distribution to the remaining six datasets.
When the base model is augmented with a {Reranker}, performance is consistently improved for {RetRobust}, {InstructRAG}, {PA-RAG}, and \alg across all benchmarks, with particularly pronounced gains on single-hop datasets such as {NQ}, {ASQA}, and {TriviaQA}.
Under this setting, \alg achieves the highest average accuracy of $69.48$, outperforming all baseline configurations.
In contrast, augmenting the base model with reasoning-aware ranking via {RankCoT} yields a different performance pattern: while its impact is limited on single-hop benchmarks, it provides the greatest improvements on {2Wiki}, which requires multi-hop reasoning over longer contexts. We attribute this degradation to {RankCoT} replacing the raw retrieved documents with compressed CoT representations, which prevents \alg from fully leveraging its selective extraction capability learned from raw contexts. We further evaluate our method using LLMEval~\cite{rau2024bergen} (Appendix~\ref{figure3}), which confirms that \alg consistently achieves superior answer accuracy and evidence faithfulness. 



\subsection{Analysis}
\label{sec:Analysis}

\paragraph{Layer-wise Distribution of Identified Neurons.}
\label{para:layerdistribution}

We select the top-3 layers with the highest density of $\mathcal{P}_{\text{irrel}}$ and $\mathcal{P}_{\text{shared}}$ as the noise filtering targets. This choice is based on the following observations from Figure~\ref{fig:neuron_where}. First, $\mathcal{P}_{\text{rel}}$ is mainly concentrated in the middle layers, peaking around Layer 15. In contrast, $\mathcal{P}_{\text{irrel}}$ and $\mathcal{P}_{\text{shared}}$ are most prominent in the later layers, especially Layers 29--31. 
Figure~\ref{fig:neuron_where} (Right) further highlights this layer-wise separation by comparing $\mathcal{P}_{\text{rel}}$ with the combined distribution of $\mathcal{P}_{\text{irrel}}$ and $\mathcal{P}_{\text{shared}}$. 
These patterns motivate our density-based selection of the top-3 layers for noise filtering.
Further analysis on the number of selected layers is provided in Appendix~\ref{appendix:numlayer_selection}.

\begin{table}[t]
    \centering
    \captionsetup{font=small}
\resizebox{\linewidth}{!}{%
\begin{tabular}{lcccccccc}
    \toprule
    \textbf{Model / Method} &
    \textbf{NQ} &
    \textbf{ASQA} &
    \textbf{TriviaQA} &
    \textbf{SCIQ} &
    \textbf{POPQA} &
    \textbf{HotpotQA} &
    \textbf{2Wiki} &
    \textbf{AVG.} \\
    \midrule

    \multicolumn{9}{l}{\textbf{\textit{Mistral-7B-Instruct-v0.2}}} \\
    RALM
    & 64.85 & 71.20 & 89.59 & 60.80 & 62.78 & 48.39 & 48.80 & 63.77 \\
    \textbf{\alg}
    & \textbf{66.47} & \textbf{72.57} & \textbf{90.04}
    & \textbf{62.50} & \textbf{64.15} & \textbf{50.46}
    & \textbf{50.65} & \textbf{65.26} \\

    \midrule

    \multicolumn{9}{l}{\textbf{\textit{Qwen-2.5-14B-Instruct}}} \\
    RALM
    & 66.33 & 72.99 & 91.29 & 59.40 & 63.72 & 52.73
    & \textbf{50.97} & 65.35 \\
    \textbf{\alg}
    & \textbf{70.42} & \textbf{75.94} & \textbf{92.49}
    & \textbf{67.00} & \textbf{66.36} & \textbf{53.42}
    & 50.32 & \textbf{67.99} \\

    \midrule

    \multicolumn{9}{l}{\textbf{\textit{Qwen3-8B}}} \\
    RALM
    & 63.41 & 68.24 & 84.44 & 54.80 & --
    & 40.00 & -- & 62.18 \\
    \textbf{\alg}
    & \textbf{65.09} & \textbf{69.62} & \textbf{88.66}
    & \textbf{55.10} & --
    & \textbf{46.60} & -- & \textbf{65.01} \\

    \midrule

    \multicolumn{9}{l}{\textbf{\textit{LLaMA-2-13B}}} \\
    RALM
    & 60.84 & 67.51 & 88.27 & 57.40 & 55.35 & 45.66 & 44.42 & 59.92 \\
    RetRobust
    & 53.23 & 61.92 & 89.80 & 58.20 & 53.24 & 40.84 & 35.17 & 56.06 \\
    \textbf{\alg}
    & \textbf{69.15} & \textbf{75.31} & \textbf{90.91}
    & \textbf{63.20} & \textbf{65.31} & \textbf{50.91}
    & \textbf{44.87} & \textbf{65.66} \\

    \bottomrule
\end{tabular}}
\caption{Comparison of accuracy across various backbone models. Best results are in \textbf{bold}.}
\label{tab:full_generator}
\end{table}
\vspace{-2mm}

\paragraph{Generalization across Generator Backbones.}
To test generalization across generators, we apply our method to {Mistral-7B-Instruct-v0.2}~\cite{jiang2023mistral7b}, {Qwen-2.5-14B-Instruct}~\cite{qwen2025qwen25technicalreport}, Qwen3-8B~\cite{yang2025qwen3} and {LLaMA-2-13B-Chat}~\cite{touvron2023llama2openfoundation}, varying architecture and scale while keeping other components fixed. 
As shown in Table~\ref{tab:full_generator}, \alg consistently outperforms the {RALM} baseline in average accuracy across various backbone models. 
The consistent performance gains across generators, together with the aligned distributional patterns observed across the analyzed backbones\footnote{We further analyze the layer-wise neuron distributions for Mistral-7B-Instruct-v0.2, Qwen-2.5-14B-Instruct, and LLaMA-2-13B-Chat, with details provided in Appendix~\ref{sec:neuron_dist_generators}.}, suggest that \alg transfers across different generators and may capture a general mechanism for improving robustness in RALMs. 

\paragraph{Irrelevant Contexts Handling.}
To verify the robustness of \alg, we evaluate its performance in two distinct scenarios: 1) where retrieved contexts contain the ground-truth answer (\textit{Relevant}), and 2) where they do not (\textit{Irrelevant}). 
As summarized in Table~\ref{tab:relevant}, \alg consistently outperforms all baseline methods across both settings. 
Notably, in the Irrelevant scenario where model robustness is most severely tested, \alg achieves the highest accuracy on all three datasets (NQ, HotpotQA, and SCIQ).
This suggests that our approach more effectively filters out retrieval noise and mitigates the risk of hallucination without the need for additional external modules. 

\begin{table}[t]
    \centering
    \captionsetup{font=small}

        \resizebox{\linewidth}{!}{%
        \begin{tabular}{lcccccccc}
            \toprule
            && \multicolumn{3}{c}{\textbf{Relevant}} && \multicolumn{3}{c}{\textbf{Irrelevant}} \\
            \cmidrule(lr){3-5} \cmidrule(lr){7-9}
            \textbf{Method}  && \textbf{NQ}  & \textbf{HotpotQA} & \textbf{SCIQ} 
            && \textbf{NQ}  & \textbf{HotpotQA} & \textbf{SCIQ} \\ 
            \midrule
            
            RALM
            && 79.16  & 73.15  & 71.08 
            && \underline{20.43}  & \underline{15.19} & \underline{17.49} \\ 
            
            InstructRAG
            && 76.87  & 74.29 & 75.04 
            && 19.54  & 11.98 & 16.62 \\ 
            
            PA-RAG
            && \underline{86.38}  & \textbf{81.96} & \underline{79.45} 
            && 20.30 & 12.32 & 12.83 \\ 
            \midrule
            
            \textbf{\alg}
            && \textbf{86.73} & \underline{81.75} & \textbf{82.50} 
            && \textbf{25.76}  & \textbf{18.45} & \textbf{20.41} \\
            \bottomrule
        \end{tabular}}
        \vspace{-2mm}
        \caption{The comparison of the accuracy of our method in scenarios with relevant (answer-present) and irrelevant (answer-absent) retrieved contexts}
        \label{tab:relevant}

\end{table}

\subsection{Ablation Study}
\label{sec:Ablation Study}
We conduct a comprehensive ablation study to examine the contribution of each design component in \alg{}. Specifically, we analyze (1) the contribution of each stage in the two-stage training framework, (2) the importance of neuron-guided layer selection, (3) the respective contributions of layer-level and neuron-level updates within the Noise Filtering Enhancement stage, and (4) sensitivity to the neuron selection budget. 
The results are summarized in Table~\ref{tab:ablation_ours}.
Additionally, we evaluate the deployment flexibility of \alg{} under retriever substitution in Table~\ref {tab:contriever_inference}. 


\paragraph{Stage-level contributions.}
We examine whether each of the two training stages independently contributes to the final performance. As shown in Table~\ref{tab:ablation_ours}, removing either the denoising stage (\emph{w/o denoising}) or the noise filtering stage (\emph{w/o NoiseFilter}) consistently degrades performance, dropping the average accuracy from 68.46 to 66.38 and 61.18, respectively. Notably, applying the denoising stage \emph{alone} (\emph{w/o NoiseFilter}: 61.18) falls below the five-dataset average of the RALM baseline (62.08), confirming that the \texttt{EOT}-based suppression signal can over-suppress useful evidence when partially relevant contexts are present. The subsequent noise filtering stage compensates by teaching the model to extract evidence from mixed-quality contexts, and the two stages thus operate as a {composed} mechanism rather than independent contributions. Furthermore, replacing this composed pipeline with standard LoRA fine-tuning on the same $\mathcal{D}_{\text{RS}}$ data (\emph{w/o denoising, NoiseFilter}: 62.33) still falls 6.13 points behind \alg, isolating the contribution of the neuron-guided framework from the distilled supervision itself.

\begin{table}[t]
    \centering
    \small
    \setlength{\tabcolsep}{3pt}
    \renewcommand{\arraystretch}{1.05}

    \resizebox{\linewidth}{!}{%
    \begin{tabular}{lcccccc}
        \toprule
        \textbf{Method} 
        & \textbf{NQ} 
        & \textbf{HotpotQA} 
        & \textbf{TriviaQA} 
        & \textbf{SCIQ} 
        & \textbf{POPQA} 
        & \textbf{Avg.} \\
        \midrule

        \textbf{\alg}
        & 69.68 & 52.25 & 92.21 & 60.80 & 67.37 & \textbf{68.46} \\
        \midrule

        \multicolumn{7}{@{}l}{\textit{Stage-level ablations}} \\
        \quad w/o denoising 
        & 67.88 & 50.37 & 91.16 & 56.60 & 65.87 & 66.38 \\

        \quad w/o NoiseFilter 
        & 61.50 & 44.82 & 87.88 & 52.10 & 59.60 & 61.18 \\

        \quad w/o denoising, NoiseFilter 
        & 62.81 & 48.28 & 88.68 & 52.40 & 59.50 & 62.33 \\
        \hdashline 
        
        \multicolumn{7}{@{}l}{\textit{Neuron-guided layer selection vs. random selection}} \\
        \quad Random layer sel. 
        & 65.31 & 45.53 & 89.54 & 55.60 & 62.66 & 63.73 \\
        \hdashline

        \multicolumn{7}{@{}l}{\textit{Component ablations within NoiseFilter}} \\
        \quad w/o NoiseFilter, Layers 
        & 61.93 & 47.10 & 88.98 & 51.90 & 58.94 & 61.77 \\

        \quad w/o NoiseFilter, Neurons 
        & 69.01 & 53.26 & 91.85 & 59.90 & 67.00 & 68.20 \\

        \quad w/o NoiseFilter, $\mathcal{D}_{\text{RS}}$ 
        & 67.53 & 51.60 & 91.14 & 58.10 & 65.82 & 66.84 \\
        \hdashline

        \multicolumn{7}{@{}l}{\textit{Sensitivity to neuron budget}} \\
        \quad Half $(50, 15, 50)$
        & 70.28 & 50.87 & 92.06 & 60.40 & 67.16 & 68.15 \\

        \quad Double $(200, 60, 200)$
        & 70.60 & 50.96 & 91.61 & 60.80 & 67.37 & 68.27 \\




        \bottomrule
    \end{tabular}}
    \vspace{-2mm}
    \caption{Ablation results of \alg across five QA datasets. Average accuracy is computed over the five datasets.}
    \label{tab:ablation_ours}
\end{table}
\vspace{-2mm}

\paragraph{Neuron-guided layer selection drives the gain.}
A natural concern is whether \alg's gains can be attributed to neuron mining at all, or whether simply fine-tuning a few layers would suffice. To isolate this, we sample neurons uniformly at random while matching the original group sizes $(|\mathcal{P}_{\text{rel}}|, |\mathcal{P}_{\text{shared}}|, |\mathcal{P}_{\text{irrel}}|) = (100, 30, 100)$, then select the top-3 layers with the highest density of these randomly sampled neurons and apply identical two-stage tuning. As reported in Table~\ref{tab:ablation_ours}, this {Random layer selection} (\textit{Random layer sel.}) reaches only 63.73 average accuracy, a 4.73-point gap from \alg (68.46). 
This confirms that the primary role of neuron mining is to provide a {localization signal}: it identifies {where} in the network adaptation should occur. Without this guidance, layer-level tuning becomes arbitrary and substantially less effective.

\paragraph{Layer-level vs. neuron-level contributions within NoiseFilter.}
Within the noise filtering stage, we update two complementary categories: (i) the top-3 layers selected by neuron density, and (ii) the identified neuron groups,
$\mathcal{P}_{\text{rel}}$, $\mathcal{P}_{\text{shared}}$, and
$\mathcal{P}_{\text{irrel}}$. 
Removing the layer-level updates (\emph{w/o NoiseFilter, Layers}) drops performance to 61.77, whereas removing only the neuron-level updates (\emph{w/o NoiseFilter, Neurons}) yields a smaller drop to 68.20. Despite this smaller gap, neuron-level updates provide a meaningful additional gain beyond layer-level selection, suggesting that fine-grained neuron-level refinement complements the broader localization provided by layer selection. This observation is consistent with prior findings that FFN computations depend on distributed parameter interactions~\citep{leng2025towards}. We also find that replacing $\mathcal{D}{_\text{RS}}$ with raw QA pairs (\emph{w/o NoiseFilter, $\mathcal{D}_{\text{RS}}$}: 66.84)  degrades performance, further confirming the value of relevant-summary supervision as the training target.

\begin{table}[t]
    \centering
    \setlength{\tabcolsep}{4.5pt}
    \renewcommand{\arraystretch}{1.10}

    \resizebox{\linewidth}{!}{%
    \begin{tabular}{lcccccc}
        \toprule
        \textbf{Method} &
        \textbf{NQ} &
        \textbf{ASQA} &
        \textbf{TriviaQA} &
        \textbf{SCIQ} &
        \textbf{HotpotQA} &
        \textbf{Avg.} \\
        \midrule

        \multicolumn{7}{l}{\textit{Contriever inference}} \\  
        \hspace{0.5em}\textbf{\alg}
                & \textbf{67.98}
                & \textbf{71.72}
                & \textbf{89.28}
                & \textbf{60.20}
                & \textbf{47.26}
                & \textbf{67.29} \\  \midrule

        \hspace{0.5em}RALM
        & 59.11 & 63.50 & 85.62 & 51.80 & 41.50 & 60.31 \\

        \hspace{0.5em}RetRobust
        & 59.67 & 59.49 & 88.90 & 56.90 & 40.30 & 61.05 \\

        \hspace{0.5em}PA-RAG
        & 62.84 & 70.04 & 86.94 & 53.50 & 44.41 & 63.55 \\

        \bottomrule
    \end{tabular}%
    }

    \caption{
    Performance under Contriever retrieval.
    For \alg{}, neurons are mined using SPLADE-v3 and reused at inference with Contriever without re-mining.
    Best results are in \textbf{bold}.
    }
    \label{tab:contriever_inference}
\end{table}



\paragraph{Sensitivity to neuron budget.}
We examine sensitivity to the number of selected neurons by halving or doubling the original budget. Both settings (68.15 and 68.27) remain within 0.31 points of the original configuration (68.46), demonstrating that \alg is robust to the precise neuron count. 


\paragraph{Retriever Flexibility.}
To assess the deployment flexibility of \alg{}, we mine context-aware neurons using SPLADE-v3 and evaluate the resulting model at inference time with a different retriever, Contriever~\citep{izacard2021unsupervised}.
While our default setting uses SPLADE-v3 for both neuron mining and inference, we replace only the inference-time retriever with Contriever, keeping the neurons mined using SPLADE-v3 fixed.
As shown in Table~\ref{tab:contriever_inference}, the SPLADE-v3-mined neurons remain effective under Contriever retrieval, with \alg{} consistently outperforming baseline methods evaluated under the same Contriever retrieval setting across all datasets.
These results suggest that the identified neurons are not tightly coupled to the retriever used during mining and can be reused across retrievers at inference time without re-mining.
Beyond inference-time retriever substitution, we further examine the sensitivity of the neuron mining procedure itself to the choice of retriever in Appendix~\ref{appendix:retriever}. 

\vspace{-1mm}
\section{Conclusion}
\vspace{-1mm}

In this paper, we introduced \alg, a neuron-guided instruction-tuning framework for robust retrieval-augmented language models (RALMs).
\alg identifies context-aware neurons via attribution analysis and uses them to guide selective adaptation at both the neuron and layer levels.
Through two-stage instruction tuning, \alg separately learns to abstain under uninformative contexts and extract relevant evidence from partially noisy contexts.
Experiments across diverse benchmarks demonstrate that \alg consistently improves robustness to retrieval noise and outperforms strong baselines.


\section*{Limitations}

Although \alg demonstrates strong robustness across diverse QA benchmarks and multiple backbone models, one limitation remains.
While our experiments cover mid-sized open-source LLMs up to the 14B scale, the effectiveness of neuron-guided instruction tuning on substantially larger models, such as 30B or 70B-scale LLMs, remains unexplored.
Since neuron-level attribution patterns may vary with model capacity and architecture, future work should examine whether the identified context-aware neurons remain stable and effective at larger scales.

\section*{Acknowledgments}
This research was supported by the Ministry of Trade, Industry and Resources (MOTIR, Korea), under the project titled ``An Unbreachable Multilayer AI-based Security Mechanism that Continuously Adapts and Evolves in Dynamic Conditions'' (Project Number: RS-2025-02653102). 
This work was supported by the Institute of Information \& Communications Technology Planning \& Evaluation (IITP) grant funded by the Korea government (MSIT) (IITP-2026)-RS-2026-25614738.
This work was supported by the AI Star Fellowship Support Program, Institute of Information \& communications Technology Planning \& Evaluation (IITP) under the Artificial Intelligence Innovation Human Resources Development (IITP-2026-RS-2026-25548323) grant funded by the Korea government(MSIT).
This work was supported by the National Research Foundation of Korea (NRF) grant funded by the Korea government (MSIT) (RS-2025-24535182 and RS-2026-25498006).
\bibliography{custom}

@article{yang2025qwen3,
  title={Qwen3 technical report},
  author={Yang, An and Li, Anfeng and Yang, Baosong and Zhang, Beichen and Hui, Binyuan and Zheng, Bo and Yu, Bowen and Gao, Chang and Huang, Chengen and Lv, Chenxu and others},
  journal={arXiv preprint arXiv:2505.09388},
  year={2025}
}

@article{wei2024instructrag,
  title={Instructrag: Instructing retrieval-augmented generation via self-synthesized rationales},
  author={Wei, Zhepei and Chen, Wei-Lin and Meng, Yu},
  journal={arXiv preprint arXiv:2406.13629},
  year={2024}
}

@inproceedings{cuconasu2024power,
  title={The power of noise: Redefining retrieval for rag systems},
  author={Cuconasu, Florin and Trappolini, Giovanni and Siciliano, Federico and Filice, Simone and Campagnano, Cesare and Maarek, Yoelle and Tonellotto, Nicola and Silvestri, Fabrizio},
  booktitle={Proceedings of the 47th International ACM SIGIR Conference on Research and Development in Information Retrieval},
  pages={719--729},
  year={2024}
}

@inproceedings{sundararajan2017axiomatic,
  title={Axiomatic attribution for deep networks},
  author={Sundararajan, Mukund and Taly, Ankur and Yan, Qiqi},
  booktitle={International conference on machine learning},
  pages={3319--3328},
  year={2017},
  organization={PMLR}
}

@inproceedings{mallen2023not,
  title={When not to trust language models: Investigating effectiveness of parametric and non-parametric memories},
  author={Mallen, Alex and Asai, Akari and Zhong, Victor and Das, Rajarshi and Khashabi, Daniel and Hajishirzi, Hannaneh},
  booktitle={Proceedings of the 61st Annual Meeting of the Association for Computational Linguistics (Volume 1: Long Papers)},
  pages={9802--9822},
  year={2023}
}

@article{ho2020constructing,
  title={Constructing a multi-hop qa dataset for comprehensive evaluation of reasoning steps},
  author={Ho, Xanh and Nguyen, Anh-Khoa Duong and Sugawara, Saku and Aizawa, Akiko},
  journal={arXiv preprint arXiv:2011.01060},
  year={2020}
}

@inproceedings{rau2024bergen,
  title={Bergen: A benchmarking library for retrieval-augmented generation},
  author={Rau, David and D{\'e}jean, Herv{\'e} and Chirkova, Nadezhda and Formal, Thibault and Wang, Shuai and Clinchant, St{\'e}phane and Nikoulina, Vassilina},
  booktitle={Findings of the Association for Computational Linguistics: EMNLP 2024},
  pages={7640--7663},
  year={2024}
}

@article{dubey2024llama,
  title={The llama 3 herd of models},
  author={Dubey, Abhimanyu and Jauhri, Abhinav and Pandey, Abhinav and Kadian, Abhishek and Al-Dahle, Ahmad and Letman, Aiesha and Mathur, Akhil and Schelten, Alan and Yang, Amy and Fan, Angela and others},
  journal={arXiv preprint arXiv:2407.21783},
  year={2024}
}

@misc{jiang2023mistral7b,
      title={Mistral 7B}, 
      author={Albert Q. Jiang and Alexandre Sablayrolles and Arthur Mensch and Chris Bamford and Devendra Singh Chaplot and Diego de las Casas and Florian Bressand and Gianna Lengyel and Guillaume Lample and Lucile Saulnier and Lélio Renard Lavaud and Marie-Anne Lachaux and Pierre Stock and Teven Le Scao and Thibaut Lavril and Thomas Wang and Timothée Lacroix and William El Sayed},
      year={2023},
      eprint={2310.06825},
      archivePrefix={arXiv},
      primaryClass={cs.CL},
      url={https://arxiv.org/abs/2310.06825}, 
}

@misc{lassance2024spladev3newbaselinessplade,
      title={SPLADE-v3: New baselines for SPLADE}, 
      author={Carlos Lassance and Hervé Déjean and Thibault Formal and Stéphane Clinchant},
      year={2024},
      eprint={2403.06789},
      archivePrefix={arXiv},
      primaryClass={cs.IR},
      url={https://arxiv.org/abs/2403.06789}, 
}

@inproceedings{yang2018hotpotqa,
  title={HotpotQA: A Dataset for Diverse, Explainable Multi-hop Question Answering},
  author={Yang, Zhilin and Qi, Peng and Zhang, Saizheng and Bengio, Yoshua and Cohen, William and Salakhutdinov, Ruslan and Manning, Christopher D},
  booktitle={Proceedings of the 2018 Conference on Empirical Methods in Natural Language Processing},
  pages={2369--2380},
  year={2018}
}

@inproceedings{welbl2017crowdsourcing,
  title={Crowdsourcing Multiple Choice Science Questions},
  author={Welbl, Johannes and Liu, Nelson F and Gardner, Matt},
  booktitle={Proceedings of the 3rd Workshop on Noisy User-generated Text},
  pages={94--106},
  year={2017}
}

@inproceedings{joshi2017triviaqa,
  title={TriviaQA: A Large Scale Distantly Supervised Challenge Dataset for Reading Comprehension},
  author={Joshi, Mandar and Choi, Eunsol and Weld, Daniel S and Zettlemoyer, Luke},
  booktitle={Proceedings of the 55th Annual Meeting of the Association for Computational Linguistics (Volume 1: Long Papers)},
  pages={1601--1611},
  year={2017}
}

@inproceedings{wu-etal-2025-pa,
    title = "{PA}-{RAG}: {RAG} Alignment via Multi-Perspective Preference Optimization",
    author = "Wu, Jiayi  and
      Cai, Hengyi  and
      Yan, Lingyong  and
      Sun, Hao  and
      Li, Xiang  and
      Wang, Shuaiqiang  and
      Yin, Dawei  and
      Gao, Ming",
    editor = "Chiruzzo, Luis  and
      Ritter, Alan  and
      Wang, Lu",
    booktitle = "Proceedings of the 2025 Conference of the Nations of the Americas Chapter of the Association for Computational Linguistics: Human Language Technologies (Volume 1: Long Papers)",
    month = apr,
    year = "2025",
    address = "Albuquerque, New Mexico",
    publisher = "Association for Computational Linguistics",
    url = "https://aclanthology.org/2025.naacl-long.459/",
    doi = "10.18653/v1/2025.naacl-long.459",
    pages = "9091--9112",
    ISBN = "979-8-89176-189-6"
}

@inproceedings{wang-etal-2023-self-knowledge,
    title = "Self-Knowledge Guided Retrieval Augmentation for Large Language Models",
    author = "Wang, Yile  and
      Li, Peng  and
      Sun, Maosong  and
      Liu, Yang",
    editor = "Bouamor, Houda  and
      Pino, Juan  and
      Bali, Kalika",
    booktitle = "Findings of the Association for Computational Linguistics: EMNLP 2023",
    month = dec,
    year = "2023",
    address = "Singapore",
    publisher = "Association for Computational Linguistics",
    url = "https://aclanthology.org/2023.findings-emnlp.691/",
    doi = "10.18653/v1/2023.findings-emnlp.691",
    pages = "10303--10315"
}

@inproceedings{vig-etal-2022-exploring,
    title = "Exploring Neural Models for Query-Focused Summarization",
    author = "Vig, Jesse  and
      Fabbri, Alexander  and
      Kryscinski, Wojciech  and
      Wu, Chien-Sheng  and
      Liu, Wenhao",
    editor = "Carpuat, Marine  and
      de Marneffe, Marie-Catherine  and
      Meza Ruiz, Ivan Vladimir",
    booktitle = "Findings of the Association for Computational Linguistics: NAACL 2022",
    month = jul,
    year = "2022",
    address = "Seattle, United States",
    publisher = "Association for Computational Linguistics",
    url = "https://aclanthology.org/2022.findings-naacl.109/",
    doi = "10.18653/v1/2022.findings-naacl.109",
    pages = "1455--1468"
}

@article{xu2023recomp,
  title={Recomp: Improving retrieval-augmented lms with compression and selective augmentation},
  author={Xu, Fangyuan and Shi, Weijia and Choi, Eunsol},
  journal={arXiv preprint arXiv:2310.04408},
  year={2023}
}

@article{shi2024ircan,
  title={Ircan: Mitigating knowledge conflicts in llm generation via identifying and reweighting context-aware neurons},
  author={Shi, Dan and Jin, Renren and Shen, Tianhao and Dong, Weilong and Wu, Xinwei and Xiong, Deyi},
  journal={Advances in Neural Information Processing Systems},
  volume={37},
  pages={4997--5024},
  year={2024}
}

@article{xu2024parenting,
  title={Parenting: Optimizing knowledge selection of retrieval-augmented language models with parameter decoupling and tailored tuning},
  author={Xu, Yongxin and Zhang, Ruizhe and Jiang, Xinke and Feng, Yujie and Xiao, Yuzhen and Ma, Xinyu and Zhu, Runchuan and Chu, Xu and Zhao, Junfeng and Wang, Yasha},
  journal={arXiv preprint arXiv:2410.10360},
  year={2024}
}

@inproceedings{wu2025rankcot,
    title = "{R}ank{C}o{T}: Refining Knowledge for Retrieval-Augmented Generation through Ranking Chain-of-Thoughts",
    author = "Wu, Mingyan  and
      Liu, Zhenghao  and
      Yan, Yukun  and
      Li, Xinze  and
      Yu, Shi  and
      Zeng, Zheni  and
      Gu, Yu  and
      Yu, Ge",
    editor = "Che, Wanxiang  and
      Nabende, Joyce  and
      Shutova, Ekaterina  and
      Pilehvar, Mohammad Taher",
    booktitle = "Proceedings of the 63rd Annual Meeting of the Association for Computational Linguistics (Volume 1: Long Papers)",
    month = jul,
    year = "2025",
    address = "Vienna, Austria",
    publisher = "Association for Computational Linguistics",
    url = "https://aclanthology.org/2025.acl-long.629/",
    pages = "12857--12874",
    ISBN = "979-8-89176-251-0"
}

@article{izacard2023atlas,
  title={Atlas: Few-shot learning with retrieval augmented language models},
  author={Izacard, Gautier and Lewis, Patrick and Lomeli, Maria and Hosseini, Lucas and Petroni, Fabio and Schick, Timo and Dwivedi-Yu, Jane and Joulin, Armand and Riedel, Sebastian and Grave, Edouard},
  journal={Journal of Machine Learning Research},
  volume={24},
  number={251},
  pages={1--43},
  year={2023}
}

@article{kwiatkowski2019natural,
  title={Natural questions: a benchmark for question answering research},
  author={Kwiatkowski, Tom and Palomaki, Jennimaria and Redfield, Olivia and Collins, Michael and Parikh, Ankur and Alberti, Chris and Epstein, Danielle and Polosukhin, Illia and Devlin, Jacob and Lee, Kenton and others},
  journal={Transactions of the Association for Computational Linguistics},
  volume={7},
  pages={453--466},
  year={2019},
  publisher={MIT Press One Rogers Street, Cambridge, MA 02142-1209, USA journals-info~…}
}

@inproceedings{stelmakh2022asqa,
    title = "{ASQA}: Factoid Questions Meet Long-Form Answers",
    author = "Stelmakh, Ivan  and
      Luan, Yi  and
      Dhingra, Bhuwan  and
      Chang, Ming-Wei",
    editor = "Goldberg, Yoav  and
      Kozareva, Zornitsa  and
      Zhang, Yue",
    booktitle = "Proceedings of the 2022 Conference on Empirical Methods in Natural Language Processing",
    month = dec,
    year = "2022",
    address = "Abu Dhabi, United Arab Emirates",
    publisher = "Association for Computational Linguistics",
    url = "https://aclanthology.org/2022.emnlp-main.566/",
    doi = "10.18653/v1/2022.emnlp-main.566",
    pages = "8273--8288"
}

@article{lewis2020retrieval,
  title={Retrieval-augmented generation for knowledge-intensive nlp tasks},
  author={Lewis, Patrick and Perez, Ethan and Piktus, Aleksandra and Petroni, Fabio and Karpukhin, Vladimir and Goyal, Naman and K{\"u}ttler, Heinrich and Lewis, Mike and Yih, Wen-tau and Rockt{\"a}schel, Tim and others},
  journal={Advances in neural information processing systems},
  volume={33},
  pages={9459--9474},
  year={2020}
}

@article{wang2024rat,
  title={Rat: Retrieval augmented thoughts elicit context-aware reasoning in long-horizon generation},
  author={Wang, Zihao and Liu, Anji and Lin, Haowei and Li, Jiaqi and Ma, Xiaojian and Liang, Yitao},
  journal={arXiv preprint arXiv:2403.05313},
  year={2024}
}

@article{huang2025survey,
  title={A survey on hallucination in large language models: Principles, taxonomy, challenges, and open questions},
  author={Huang, Lei and Yu, Weijiang and Ma, Weitao and Zhong, Weihong and Feng, Zhangyin and Wang, Haotian and Chen, Qianglong and Peng, Weihua and Feng, Xiaocheng and Qin, Bing and others},
  journal={ACM Transactions on Information Systems},
  volume={43},
  number={2},
  pages={1--55},
  year={2025},
  publisher={ACM New York, NY}
}

@inproceedings{lin2025llm,
  title={LLM whisperer: An inconspicuous attack to bias LLM responses},
  author={Lin, Weiran and Gerchanovsky, Anna and Akgul, Omer and Bauer, Lujo and Fredrikson, Matt and Wang, Zifan},
  booktitle={Proceedings of the 2025 CHI Conference on Human Factors in Computing Systems},
  pages={1--24},
  year={2025}
}

@inproceedings{yu-etal-2022-retrieval,
    title = "Retrieval Augmentation for Commonsense Reasoning: A Unified Approach",
    author = "Yu, Wenhao  and
      Zhu, Chenguang  and
      Zhang, Zhihan  and
      Wang, Shuohang  and
      Zhang, Zhuosheng  and
      Fang, Yuwei  and
      Jiang, Meng",
    editor = "Goldberg, Yoav  and
      Kozareva, Zornitsa  and
      Zhang, Yue",
    booktitle = "Proceedings of the 2022 Conference on Empirical Methods in Natural Language Processing",
    month = dec,
    year = "2022",
    address = "Abu Dhabi, United Arab Emirates",
    publisher = "Association for Computational Linguistics",
    url = "https://aclanthology.org/2022.emnlp-main.294/",
    doi = "10.18653/v1/2022.emnlp-main.294",
    pages = "4364--4377"
}

@inproceedings{rau-etal-2024-bergen,
    title = "{BERGEN}: A Benchmarking Library for Retrieval-Augmented Generation",
    author = "Rau, David  and
      D{\'e}jean, Herv{\'e}  and
      Chirkova, Nadezhda  and
      Formal, Thibault  and
      Wang, Shuai  and
      Clinchant, St{\'e}phane  and
      Nikoulina, Vassilina",
    editor = "Al-Onaizan, Yaser  and
      Bansal, Mohit  and
      Chen, Yun-Nung",
    booktitle = "Findings of the Association for Computational Linguistics: EMNLP 2024",
    month = nov,
    year = "2024",
    address = "Miami, Florida, USA",
    publisher = "Association for Computational Linguistics",
    url = "https://aclanthology.org/2024.findings-emnlp.449/",
    doi = "10.18653/v1/2024.findings-emnlp.449",
    pages = "7640--7663"
}

@inproceedings{lin2023ra,
  title={Ra-dit: Retrieval-augmented dual instruction tuning},
  author={Lin, Xi Victoria and Chen, Xilun and Chen, Mingda and Shi, Weijia and Lomeli, Maria and James, Richard and Rodriguez, Pedro and Kahn, Jacob and Szilvasy, Gergely and Lewis, Mike and others},
  booktitle={The Twelfth International Conference on Learning Representations},
  year={2023}
}

@article{gao2023retrieval,
  title={Retrieval-augmented generation for large language models: A survey},
  author={Gao, Yunfan and Xiong, Yun and Gao, Xinyu and Jia, Kangxiang and Pan, Jinliu and Bi, Yuxi and Sun, Jiawei and Wang, Haofen},
  year={2023}
}

@inproceedings{guu2020retrieval,
  title={Retrieval augmented language model pre-training},
  author={Guu, Kelvin and Lee, Kenton and Tung, Zora and Pasupat, Panupong and Chang, Mingwei},
  booktitle={International conference on machine learning},
  pages={3929--3938},
  year={2020},
  organization={PMLR}
}

@article{jiang2024llms,
  title={Do llms dream of elephants (when told not to)? latent concept association and associative memory in transformers},
  author={Jiang, Yibo and Rajendran, Goutham and Ravikumar, Pradeep and Aragam, Bryon},
  journal={Advances in Neural Information Processing Systems},
  volume={37},
  pages={67712--67757},
  year={2024}
}

@inproceedings{wang-yu-2025-iquest,
    title = "i{QUEST}: An Iterative Question-Guided Framework for Knowledge Base Question Answering",
    author = "Wang, Shuai  and
      Yu, Yinan",
    editor = "Che, Wanxiang  and
      Nabende, Joyce  and
      Shutova, Ekaterina  and
      Pilehvar, Mohammad Taher",
    booktitle = "Proceedings of the 63rd Annual Meeting of the Association for Computational Linguistics (Volume 1: Long Papers)",
    month = jul,
    year = "2025",
    address = "Vienna, Austria",
    publisher = "Association for Computational Linguistics",
    url = "https://aclanthology.org/2025.acl-long.760/",
    pages = "15616--15628",
    ISBN = "979-8-89176-251-0"
}

@inproceedings{fan2024survey,
  title={A survey on rag meeting llms: Towards retrieval-augmented large language models},
  author={Fan, Wenqi and Ding, Yujuan and Ning, Liangbo and Wang, Shijie and Li, Hengyun and Yin, Dawei and Chua, Tat-Seng and Li, Qing},
  booktitle={Proceedings of the 30th ACM SIGKDD conference on knowledge discovery and data mining},
  pages={6491--6501},
  year={2024}
}

@inproceedings{trivedi2023interleaving,
  title={Interleaving Retrieval with Chain-of-Thought Reasoning for Knowledge-Intensive Multi-Step Questions},
  author={Trivedi, Harsh and Balasubramanian, Niranjan and Khot, Tushar and Sabharwal, Ashish},
  booktitle={Proceedings of the 61st Annual Meeting of the Association for Computational Linguistics (Volume 1: Long Papers)},
  pages={10014--10037},
  year={2023}
}

@inproceedings{chen2024benchmarking,
  title={Benchmarking large language models in retrieval-augmented generation},
  author={Chen, Jiawei and Lin, Hongyu and Han, Xianpei and Sun, Le},
  booktitle={Proceedings of the AAAI Conference on Artificial Intelligence},
  volume={38},
  number={16},
  pages={17754--17762},
  year={2024}
}

@inproceedings{yoran2024makingretrievalaugmentedlanguagemodels,
  title={Making Retrieval-Augmented Language Models Robust to Irrelevant Context},
  author={Yoran, Ori and Wolfson, Tomer and Ram, Ori and Berant, Jonathan},
  booktitle={The Twelfth International Conference on Learning Representations (ICLR)},
  year ={2024}
}

@inproceedings{xu2025let,
  title={Let’s focus on neuron: Neuron-level supervised fine-tuning for large language model},
  author={Xu, Haoyun and Zhan, Runzhe and Ma, Yingpeng and Wong, Derek F and Chao, Lidia S},
  booktitle={Proceedings of the 31st International Conference on Computational Linguistics},
  pages={9393--9406},
  year={2025}
}

@inproceedings{glass-etal-2022-re2g,
    title = "{R}e2{G}: Retrieve, Rerank, Generate",
    author = "Glass, Michael  and
      Rossiello, Gaetano  and
      Chowdhury, Md Faisal Mahbub  and
      Naik, Ankita  and
      Cai, Pengshan  and
      Gliozzo, Alfio",
    editor = "Carpuat, Marine  and
      de Marneffe, Marie-Catherine  and
      Meza Ruiz, Ivan Vladimir",
    booktitle = "Proceedings of the 2022 Conference of the North American Chapter of the Association for Computational Linguistics: Human Language Technologies",
    month = jul,
    year = "2022",
    address = "Seattle, United States",
    publisher = "Association for Computational Linguistics",
    url = "https://aclanthology.org/2022.naacl-main.194/",
    doi = "10.18653/v1/2022.naacl-main.194",
    pages = "2701--2715"
}

@inproceedings{leng2025towards,
  title={Towards understanding multi-task learning (generalization) of llms via detecting and exploring task-specific neurons},
  author={Leng, Yongqi and Xiong, Deyi},
  booktitle={Proceedings of the 31st International Conference on Computational Linguistics},
  pages={2969--2987},
  year={2025}
}

@article{li2022lazy,
  title={The lazy neuron phenomenon: On emergence of activation sparsity in transformers},
  author={Li, Zonglin and You, Chong and Bhojanapalli, Srinadh and Li, Daliang and Rawat, Ankit Singh and Reddi, Sashank J and Ye, Ke and Chern, Felix and Yu, Felix and Guo, Ruiqi and others},
  journal={arXiv preprint arXiv:2210.06313},
  year={2022}
}

@inproceedings{kim-etal-2024-solar,
    title = "{SOLAR} 10.7{B}: Scaling Large Language Models with Simple yet Effective Depth Up-Scaling",
    author = "Kim, Sanghoon  and
      Kim, Dahyun  and
      Park, Chanjun  and
      Lee, Wonsung  and
      Song, Wonho  and
      Kim, Yunsu  and
      Kim, Hyeonwoo  and
      Kim, Yungi  and
      Lee, Hyeonju  and
      Kim, Jihoo  and
      Ahn, Changbae  and
      Yang, Seonghoon  and
      Lee, Sukyung  and
      Park, Hyunbyung  and
      Gim, Gyoungjin  and
      Cha, Mikyoung  and
      Lee, Hwalsuk  and
      Kim, Sunghun",
    editor = "Yang, Yi  and
      Davani, Aida  and
      Sil, Avi  and
      Kumar, Anoop",
    booktitle = "Proceedings of the 2024 Conference of the North American Chapter of the Association for Computational Linguistics: Human Language Technologies (Volume 6: Industry Track)",
    month = jun,
    year = "2024",
    address = "Mexico City, Mexico",
    publisher = "Association for Computational Linguistics",
    url = "https://aclanthology.org/2024.naacl-industry.3/",
    doi = "10.18653/v1/2024.naacl-industry.3",
    pages = "23--35"
}

@misc{qwen2025qwen25technicalreport,
      title={Qwen2.5 Technical Report}, 
      author={Qwen and : and An Yang and Baosong Yang and Beichen Zhang and Binyuan Hui and Bo Zheng and Bowen Yu and Chengyuan Li and Dayiheng Liu and Fei Huang and Haoran Wei and Huan Lin and Jian Yang and Jianhong Tu and Jianwei Zhang and Jianxin Yang and Jiaxi Yang and Jingren Zhou and Junyang Lin and Kai Dang and Keming Lu and Keqin Bao and Kexin Yang and Le Yu and Mei Li and Mingfeng Xue and Pei Zhang and Qin Zhu and Rui Men and Runji Lin and Tianhao Li and Tianyi Tang and Tingyu Xia and Xingzhang Ren and Xuancheng Ren and Yang Fan and Yang Su and Yichang Zhang and Yu Wan and Yuqiong Liu and Zeyu Cui and Zhenru Zhang and Zihan Qiu},
      year={2025},
      eprint={2412.15115},
      archivePrefix={arXiv},
      primaryClass={cs.CL},
      url={https://arxiv.org/abs/2412.15115}, 
}

@misc{touvron2023llama2openfoundation,
      title={Llama 2: Open Foundation and Fine-Tuned Chat Models}, 
      author={Hugo Touvron and Louis Martin and Kevin Stone and Peter Albert and Amjad Almahairi and Yasmine Babaei and Nikolay Bashlykov and Soumya Batra and Prajjwal Bhargava and Shruti Bhosale and Dan Bikel and Lukas Blecher and Cristian Canton Ferrer and Moya Chen and Guillem Cucurull and David Esiobu and Jude Fernandes and Jeremy Fu and Wenyin Fu and Brian Fuller and Cynthia Gao and Vedanuj Goswami and Naman Goyal and Anthony Hartshorn and Saghar Hosseini and Rui Hou and Hakan Inan and Marcin Kardas and Viktor Kerkez and Madian Khabsa and Isabel Kloumann and Artem Korenev and Punit Singh Koura and Marie-Anne Lachaux and Thibaut Lavril and Jenya Lee and Diana Liskovich and Yinghai Lu and Yuning Mao and Xavier Martinet and Todor Mihaylov and Pushkar Mishra and Igor Molybog and Yixin Nie and Andrew Poulton and Jeremy Reizenstein and Rashi Rungta and Kalyan Saladi and Alan Schelten and Ruan Silva and Eric Michael Smith and Ranjan Subramanian and Xiaoqing Ellen Tan and Binh Tang and Ross Taylor and Adina Williams and Jian Xiang Kuan and Puxin Xu and Zheng Yan and Iliyan Zarov and Yuchen Zhang and Angela Fan and Melanie Kambadur and Sharan Narang and Aurelien Rodriguez and Robert Stojnic and Sergey Edunov and Thomas Scialom},
      year={2023},
      eprint={2307.09288},
      archivePrefix={arXiv},
      primaryClass={cs.CL},
      url={https://arxiv.org/abs/2307.09288}, 
}

@misc{dinan2019wizardwikipediaknowledgepoweredconversational,
      title={Wizard of Wikipedia: Knowledge-Powered Conversational agents}, 
      author={Emily Dinan and Stephen Roller and Kurt Shuster and Angela Fan and Michael Auli and Jason Weston},
      year={2019},
      eprint={1811.01241},
      archivePrefix={arXiv},
      primaryClass={cs.CL},
      url={https://arxiv.org/abs/1811.01241}, 
}

@misc{fan2019eli5longformquestion,
      title={ELI5: Long Form Question Answering}, 
      author={Angela Fan and Yacine Jernite and Ethan Perez and David Grangier and Jason Weston and Michael Auli},
      year={2019},
      eprint={1907.09190},
      archivePrefix={arXiv},
      primaryClass={cs.CL},
      url={https://arxiv.org/abs/1907.09190}, 
}

@misc{thorne2018feverlargescaledatasetfact,
      title={FEVER: a large-scale dataset for Fact Extraction and VERification}, 
      author={James Thorne and Andreas Vlachos and Christos Christodoulopoulos and Arpit Mittal},
      year={2018},
      eprint={1803.05355},
      archivePrefix={arXiv},
      primaryClass={cs.CL},
      url={https://arxiv.org/abs/1803.05355}, 
}

@misc{izacard2021contriever,
      title={Unsupervised Dense Information Retrieval with Contrastive Learning}, 
      author={Gautier Izacard and Mathilde Caron and Lucas Hosseini and Sebastian Riedel and Piotr Bojanowski and Armand Joulin and Edouard Grave},
      year={2021},
      url = {https://arxiv.org/abs/2112.09118},
      doi = {10.48550/ARXIV.2112.09118},
}

@article{izacard2021unsupervised,
  title={Unsupervised dense information retrieval with contrastive learning},
  author={Izacard, Gautier and Caron, Mathilde and Hosseini, Lucas and Riedel, Sebastian and Bojanowski, Piotr and Joulin, Armand and Grave, Edouard},
  journal={arXiv preprint arXiv:2112.09118},
  year={2021}
}

\appendix

\newcolumntype{b}{>{\columncolor{Beige}}c}

\section{Neuron Attribution and Data Construction}
\label{data_attr}

\subsection{Data for Neuron Attribution}

To compute neuron attribution scores, we adopt the Integrated Gradients (IG) formulation~\cite{sundararajan2017axiomatic} and construct a dedicated binary decision dataset tailored for attribution analysis.
The dataset is designed to probe neuron behaviors under contrasting retrieval conditions by explicitly controlling the relevance of the provided context.

\paragraph{Data Construction for Neuron Attribution}
\label{data_attr}
We employ a standard RAG pipeline with the SPLADE-v3~\cite{lassance2024spladev3newbaselinessplade} retriever over the \textsc{HotpotQA} dataset~\cite{yang2018hotpotqa}.
For each query, SPLADE-v3 assigns sparse lexical similarity scores to candidate documents.
Following the triplet construction in Section~\ref{method}, we construct 389 attribution instances
$(q_i, d_i^{\text{rel}}, d_i^{\text{irrel}})$, where $i$ indexes the attribution instances.
The relevant context $d_i^{\text{rel}}$ is selected as the top-ranked retrieved context for each query, while the irrelevant context $d_i^{\text{irrel}}$ is selected from zero-scoring retrieved contexts.
In cases where multiple contexts receive zero similarity scores, instances are selected according to their retrieval order.
The retrieved contexts are used in their raw form without additional filtering or summarization, and all attribution instances are constructed through deterministic preprocessing using Python scripts.
We then compute neuron attribution scores for relevant and irrelevant contexts separately and aggregate the selected neurons across instances, yielding the initial neuron candidate sets $\tilde{\mathcal{P}}_{\text{rel}}$ and $\tilde{\mathcal{P}}_{\text{irrel}}$.

\paragraph{Binary Attribution Task Formulation}
Each instance is formulated as a forced binary classification task.
Specifically, the model is provided with a single combined input consisting of a question, a retrieved context, and a proposed answer.
The model is instructed to determine whether the proposed answer can be derived solely from the given context by choosing between two predefined options: \texttt{YES} or \texttt{NO}.
This formulation enforces a strict binary decision and eliminates ambiguity in the output space, enabling stable and interpretable attribution analysis.
An illustrative example of the attribution dataset format is provided in Table~\ref{tb:ig_dataset_examples}.

Attribution scores are computed with respect to the loss associated with predicting the correct binary label.
During IG computation, the model is restricted to inferring answer correctness based solely on the provided context, without access to any external knowledge.
The resulting IG scores quantify the contribution of individual neurons to the model’s binary decision and are subsequently used to identify neurons with distinct functional roles.

\subsection{Examples of the Constructed Datasets}

We present qualitative examples of the two instruction-tuning datasets constructed in our framework.
These examples illustrate how the datasets provide complementary supervision:
(i) an \emph{irrelevant-context denoising dataset} encourages irrelevant-context suppression through EOT generation, while
(ii) a \emph{relevant summary dataset} $\mathcal{D}_{RS}$ promotes selective evidence integration by distilling only query-relevant information rather than performing generic document summarization.

\paragraph{Irrelevant Context Denoising Dataset.}
For the denoising stage, we construct training instances by selecting five retrieved documents that are irrelevant to the input query.
Given these irrelevant contexts, the model is explicitly instructed to identify the absence of query-relevant information and emit an End-of-Text (EOT) token.
This dataset is designed to impose a hard constraint on neurons associated with irrelevant contexts, effectively decoupling their activations from the generation process.
As a result, the model learns to terminate generation when only non-informative or misleading evidence is provided.
A representative example of this denoising dataset is shown in Table~\ref{tb:rs_example1}.

\paragraph{Relevant Summary Dataset $\mathcal{D}_{RS}$.}
The second dataset focuses on reinforcing the extraction of query-relevant evidence.
For each query, we retrieve five candidate documents and use an external LLM to construct a concise relevant summary that integrates only the information semantically aligned with the query.
For example, given the query \textit{``Who was once considered the best kickboxer in the world, however he has been involved in a number of controversies relating to his unsportsmanlike conduct and crimes of violence outside of the ring?''}, the generated summary identifies \textit{Badr Hari} as the target entity.
Among the retrieved documents, only the passage about Badr Hari contains both the required biographical information, such as \textit{``Moroccan-Dutch super heavyweight kickboxer''} and \textit{``former K-1 Heavyweight champion''}, and contextual cues about unsportsmanlike conduct and violent incidents.
Documents about unrelated athletes, such as \textit{Rau’shee Warren}, \textit{Yamaguchi Falcão}, and \textit{Rafael Carvalho}, are therefore excluded from the summary.
An illustrative example of the constructed relevant summary is provided in Table~\ref{tb:rs_example2}.


\section{Training and Implementation Details}
\label{appendix:training_details}
\subsection{Prompts}
We provide the prompts used for Denoising Irrelevant Context-Aware Neuron-level Parameters and Context-Aware Neuron-level Parameters Robustness Enforcing Fine-tuning in Section~\ref{prompt_tuning}.
\begin{tcolorbox}[
  title={Prompt for Denoising Context-Aware Neuron-level Parameters},
  colback=white,
  colframe=black,
  colbacktitle=black,
  coltitle=white,
  fonttitle=\bfseries\small,  
  boxrule=0.6pt,
  arc=2mm,
  left=1mm,right=1mm,top=1.5mm,bottom=1.5mm,
  enhanced
]
\footnotesize                
\texttt{<|begin\_of\_text|><|start\_header\_id|>user\allowbreak<|end\_header\_id|>} \\ \\
Given a document and a query, reason step by step to identify only the parts of the document that are directly relevant to the query, and provide a concise summary of those relevant parts. \\ \\
Background:\\ 
Document 1: \{document 1\}\\ 
Document 2: \{document 2\}\\ 
Document 3: \{document 3\}\\ 
Document 4: \{document 4\}\\ 
Document 5: \{document 5\}\\ \\ 
Question: \{question\}\texttt{<|eot\_id|><|start\_header\_id|>\allowbreak assistant<|end\_header\_id|>}\texttt{<|eot\_id|>}
\end{tcolorbox}
\begin{tcolorbox}[
  title={Prompt for Context-Aware Neuron-level Parameters Robustness Enforcing Fine-tuning},
  colback=white,
  colframe=black,
  colbacktitle=black,
  coltitle=white,
  fonttitle=\bfseries\small,  
  boxrule=0.6pt,
  arc=2mm,
  left=1mm,right=1mm,top=1.5mm,bottom=1.5mm,
  enhanced
]
\footnotesize                
\texttt{<|begin\_of\_text|><|start\_header\_id|>user\allowbreak<|end\_header\_id|>} \\ \\
Given a document and a query, reason step by step to identify only the parts of the document that are directly relevant to the query, and provide a concise summary of those relevant parts. \\ \\
Background:\\ 
Document 1: \{document 1\}\\ 
Document 2: \{document 2\}\\ 
Document 3: \{document 3\}\\ 
Document 4: \{document 4\}\\ 
Document 5: \{document 5\}\\ \\ 
Question: \{question\}\texttt{<|eot\_id|><|start\_header\_id|>\allowbreak assistant<|end\_header\_id|>}\\ \\
\{relevant summary\}\texttt{<|eot\_id|>}
\end{tcolorbox}
\begin{tcolorbox}[
  title={Prompt for Open-Domain Question Answering},
  colback=white,
  colframe=black,
  colbacktitle=black,
  coltitle=white,
  fonttitle=\bfseries\small,  
  boxrule=0.6pt,
  arc=2mm,
  left=1mm,right=1mm,top=1.5mm,bottom=1.5mm,
  enhanced
]
\footnotesize                
\texttt{<|begin\_of\_text|><|start\_header\_id|>system\allowbreak<|end\_header\_id|>} \\ \\ 
You are a helpful assistant. Your task is to extract relevant information from the provided documents\\ 
and answer questions as briefly as possible.\texttt{<|eot\_id|>} \\
\texttt{<|start\_header\_id|>user<|end\_header\_id|>} \\ \\ 
Background:\\ 
Document 1: \{document 1\}\\ 
Document 2: \{document 2\}\\ 
Document 3: \{document 3\}\\ 
Document 4: \{document 4\}\\ 
Document 5: \{document 5\}\\ \\ 
Question: \{question\}\texttt{<|eot\_id|><|start\_header\_id|>\allowbreak assistant<|end\_header\_id|>}
\end{tcolorbox}
\begin{tcolorbox}[
  title={Prompt for Relevant Summary Data Construction},
  colback=white,
  colframe=black,
  colbacktitle=black,
  coltitle=white,
  fonttitle=\bfseries\small,  
  boxrule=0.6pt,
  arc=2mm,
  left=1mm,right=1mm,top=1.5mm,bottom=1.5mm,
  enhanced
]
\footnotesize                
Below is a document.\\ \\ 
Your task is to find and concisely summarize only the parts of the document that are directly relevant to the given query.\\ \\ 
- Do not summarize the entire document.\\ 
- Exclude any information that is not related to the query.\\ 
- Focus only on the key points that are most relevant to the query.\\ \\ 
Query: \{query\}\\ \\ 
Document:\\ \{document\}\\ \\ 
{[}Relevant Summary{]}\\ 
(For this section, reason step-by-step in a Chain-of-Thought (CoT) manner to identify the relevant information. Show your thinking process as you determine which parts of the document are relevant to the query. Then, Summarize only the information directly related to the query based on your reasoning. If nothing is relevant, leave this section blank.)
\label{data_construct}
\end{tcolorbox}

\subsection{Implementation Details}
\label{sec:exp_setup}
All experiments are conducted using the generator LLM as {LLaMA-3-8B-Instruct}~\cite{dubey2024llama}, and training is performed on a single NVIDIA H200 GPU with 141GB memory. 
During data construction of the $\mathcal{D}_{\text{RS}}$, the external LLM produces 142-token summaries grounded exclusively in the provided input documents, with zero intrinsic knowledge usage enforced via prompting, thereby ensuring document-grounded generation and mitigating hallucination.
Our two-stage instruction tuning both use the AdamW optimizer, with the denoising stage trained at a learning rate of $1\times10^{-5}$ for one epoch and the noise filtering stage trained at a learning rate of $2\times10^{-5}$ for two epochs, using a batch size of 4 throughout.

\begin{figure*}[t]
    \centering
    \includegraphics[width=1.0\textwidth]{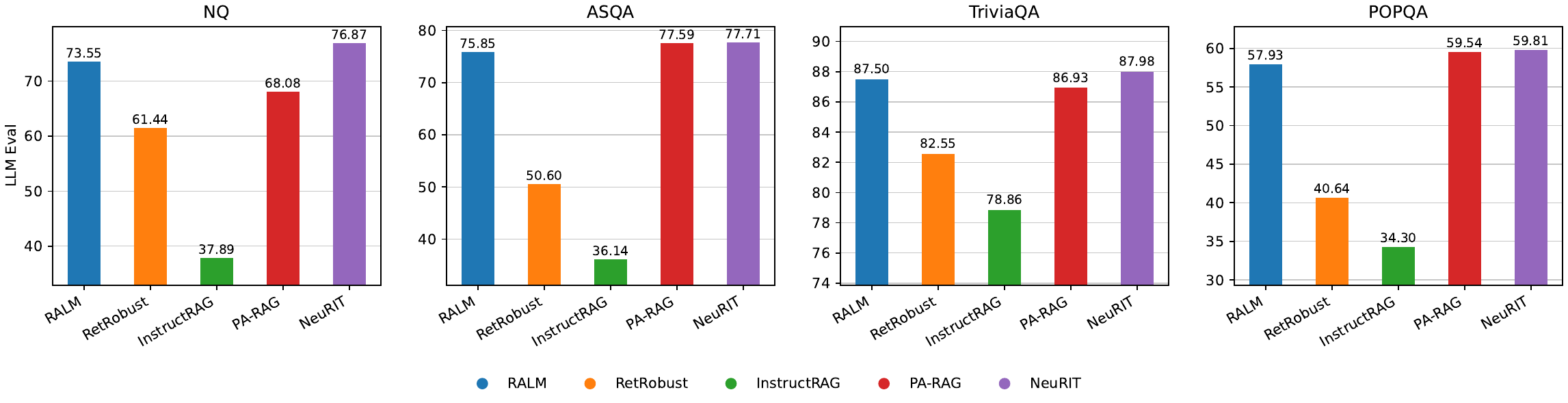}
    \caption{Performance comparison of five RAG-based methods on four QA benchmarks, evaluated using an LLM-based metric.}
    \label{fig:llm_eval_datasets}
\end{figure*}

\subsection{Neuron Mining and Selection Details}
\label{sec:neuron_selection_hyperparams}

We report the details of neuron mining and selection.
For neuron attribution, we use 389 \textsc{HotpotQA} training instances for each context type, as described in Appendix~\ref{data_attr}.
This follows the few-hundred-sample scale used in prior attribution-based context-aware neuron analysis~\cite{shi2024ircan}, while keeping neuron-level Integrated Gradients feasible on a single NVIDIA H200 GPU.
Initial mining yields 7,380 relevant and 7,036 irrelevant candidate neurons, with 6,240 overlapping neurons; removing the overlap leaves 1,140 relevant-specific and 796 irrelevant-specific candidates.
We set a compact selection budget at approximately the top 10\% scale, motivated by prior representative-neuron selection practice~\cite{shi2024ircan}.
Under our frequency-based aggregation procedure, the occurrence threshold $k=130$ yields 100 relevant and 100 irrelevant neurons, and we use the 30 most frequent overlapping neurons as $\mathcal{P}_{\text{shared}}$.

\subsection{Neuron-Guided Adaptation Details}
\label{sec:neuron_adaptation_details}

We employ stage-specific gradient masking and layer adaptation based on the neuron groups identified during neuron mining.
Our implementation of neuron-level gradient masking follows \citet{xu2025let}.
\paragraph{Denoising Stage.}
Using the irrelevant-context dataset, we target the irrelevant-context neuron set $\mathcal{P}_{\text{irrel}}$ and apply gradient masks to the FFN \texttt{up\_proj}, \texttt{down\_proj}, and \texttt{gate\_proj} projections.
This restricts parameter updates to neurons associated with irrelevant-context processing.
The model is trained to generate an End-of-Text (EOT) token for irrelevant-context inputs, thereby learning to abstain when the retrieved context is uninformative.
\paragraph{Noise-Filtering Stage.}
Using the relevant-summary dataset, we apply neuron-level gradient masks to the FFN \texttt{up\_proj} and \texttt{down\_proj} projections for all three neuron groups, $\mathcal{P}_{\text{rel}}$, $\mathcal{P}_{\text{shared}}$, and $\mathcal{P}_{\text{irrel}}$.
We additionally fully fine-tune the top-3 layers with the highest density of $\mathcal{P}_{\text{irrel}}$ and $\mathcal{P}_{\text{shared}}$.
Together, these neuron- and layer-level updates enable the model to selectively extract query-relevant evidence from partially noisy retrieved contexts.

\subsection{Detailed Computational Cost Analysis}
\noindent
\begin{minipage}{\columnwidth}
    \centering
    \small
    \setlength{\tabcolsep}{4pt}
    \renewcommand{\arraystretch}{1.05}

    \begin{tabularx}{\columnwidth}{@{}Xr@{}}
        \toprule
        \textbf{Pipeline Component} & \textbf{Time} \\
        \midrule
        Attribution data construction using the retriever (\eg SPLADE-v3)
        & $\sim$5.0 h \\

        Integrated Gradients computation over all neurons
        for 389 samples
        & $\sim$7.5 h \\

        Denoising stage
        & $\sim$1.5 h \\

        Noise Filtering Enhancement stage
        & $\sim$3.0 h \\
        \midrule


        {Full pipeline}
        & {$\sim$17.0 h} \\
        \bottomrule
    \end{tabularx}

\captionof{table}{Computation time of each component in the \alg{} pipeline on a single NVIDIA H200 GPU.}
\vspace{1mm}
    \label{tab:wallclock_time}
\end{minipage}

\noindent We further report the computation time of each component in the \alg{} pipeline on a single NVIDIA H200 GPU. As shown in Table~\ref{tab:wallclock_time}, the full pipeline requires approximately 17 hours, of which the two instruction-tuning stages account for about 4.5 hours.

\section{Evaluation Details}

\subsection{Evaluation with Match Metric}

All evaluations are conducted using the BERGEN framework~\cite{rau2024bergen} under a retrieval-augmented generation (RAG) setting.
We evaluate our method on the official test splits of the KILT-HotpotQA~\cite{yang2018hotpotqa}, and assess generalization across multiple benchmarks, including KILT-NQ~\cite{kwiatkowski2019natural}, ASQA~\cite{stelmakh2022asqa}, KILT-TriviaQA~\cite{joshi2017triviaqa}, SCIQ~\cite{welbl2017crowdsourcing}, POPQA~\cite{mallen2023not} and 2WikiMultiHopQA~\cite{ho2020constructing}.

No evaluation instances are used during training or neuron attribution computation.
For all experiments, we fix the retriever to SPLADE-v3\cite{lassance2024spladev3newbaselinessplade} to ensure consistent retrieval behavior across methods.
Given an input question, the retriever first retrieves the top-50 candidate documents based on sparse lexical similarity, from which the top-5 documents are provided to the generator as contextual input.
To ensure fair comparison, a unified prompt template is applied across all models and datasets.
Model performance is evaluated using the \textit{Match} metric provided by the BERGEN\cite{rau-etal-2024-bergen} framework, which measures exact correspondence between the generated response and the ground-truth answer.
We conduct all match-based evaluations on a single NVIDIA H200 GPU with a batch size of 256 across all benchmarks.

\subsection{LLM-based Evaluation Protocol}
\label{figure3}
In addition to exact match evaluation, we conduct LLM-based evaluations using the BERGEN framework to assess answer quality beyond strict string matching.
LLM-based evaluation is performed on four representative QA benchmarks: NQ, ASQA, TriviaQA, and POPQA.
For these datasets, we evaluate our method and baseline approaches on the development splits.

Following the BERGEN evaluation protocol, we use \texttt{SOLAR-10.7B}\cite{kim-etal-2024-solar} as the evaluation model, deployed via vLLM.
The LLM evaluator is configured with a batch size of 128.
All evaluated methods use identical retrieved contexts and prompts to ensure a controlled comparison.

As shown in Figure~\ref{fig:llm_eval_datasets}, \alg achieves the best LLM-based evaluation scores on all four benchmarks, indicating that the improvements observed under match evaluation are also reflected in answer quality judged by an LLM evaluator.

\section{Additional Analysis}
\subsection{Sensitivity to the Number of Selected Layers}
\label{appendix:numlayer_selection}
\noindent
\begin{minipage}{\columnwidth}
    \centering
    \small
    \setlength{\tabcolsep}{4pt}
    \renewcommand{\arraystretch}{1.05}

    \begin{tabularx}{\columnwidth}{@{}Xcccc@{}}
        \toprule
        \textbf{Method}
        & \textbf{NQ}
        & \textbf{ASQA}
        & \textbf{SCIQ}
        & \textbf{HotpotQA} \\
        \midrule

        RALM
        & 62.21 & 67.82 & 53.30 & 45.44 \\

        Top-2 Layers
        & 68.31 & 74.26 & 57.50 & 51.89 \\

        Top-4 Layers
        & 68.98 & 75.08 & 58.00 & 51.64 \\

        \textbf{Top-3 Layers (Ours)}
        & \textbf{69.68}
        & \textbf{75.31}
        & \textbf{60.80}
        & \textbf{52.25} \\
        \bottomrule
    \end{tabularx}

    \captionof{table}{Sensitivity analysis on the number of selected layers.}
    \vspace{1mm}
    \label{tab:num_layers}
\end{minipage}

\noindent In \alg{}, the top-3 layers with the highest density of $\mathcal{P}_{\text{irrel}}$ and $\mathcal{P}_{\text{shared}}$ are used in the Noise Filtering Enhancement stage. We analyze the effect of the number of selected layers to justify this choice. As shown in Table~\ref{tab:num_layers}, all tested settings outperform the RALM baseline, while selecting the top-3 layers achieves the best overall performance. These results justify our choice of the top-3 layers as the default configuration.

\subsection{Neuron Distribution across Generators}
\label{sec:neuron_dist_generators}

\begin{figure*}[t]
    \centering

    \textbf{(a) Mistral-7B-Instruct-v0.2}\\[-0.2em]
    \includegraphics[width=0.8\textwidth]{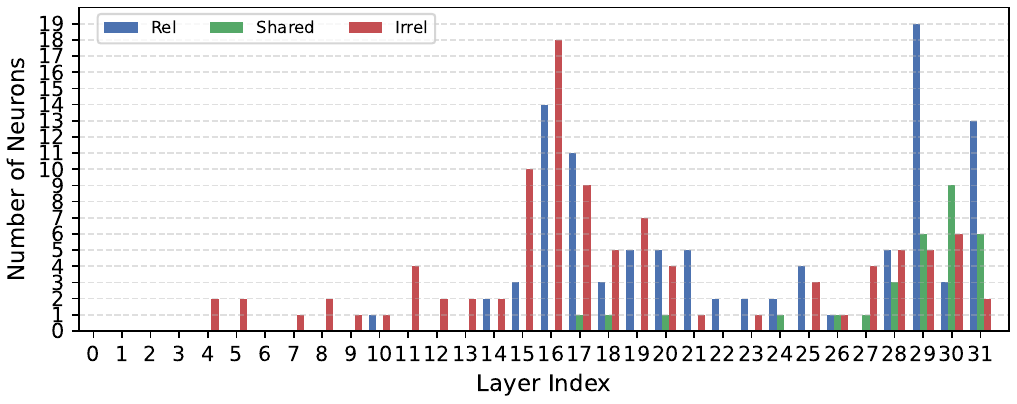}
    \vspace{0.5em}

    \textbf{(b) Qwen2.5-14B-Instruct}\\[-0.2em]
    \includegraphics[width=0.8\textwidth]{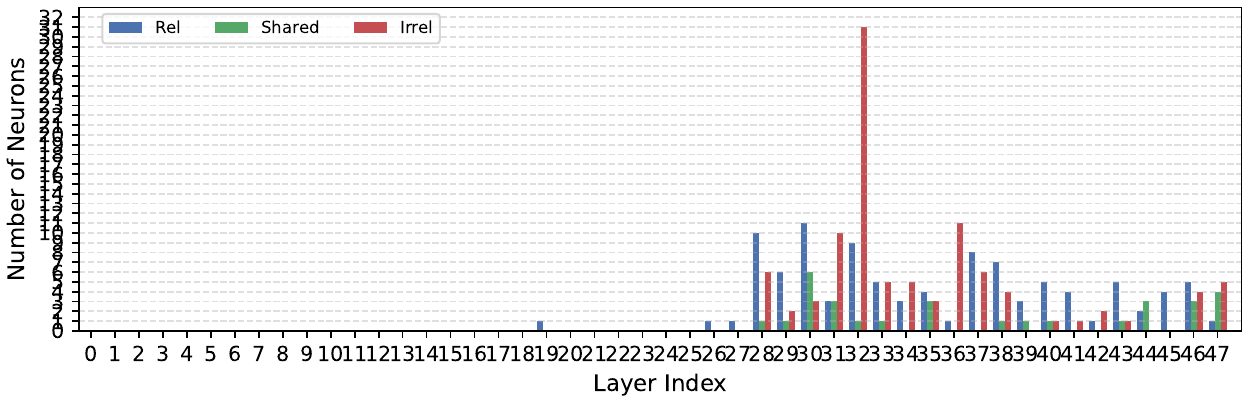}
    \vspace{0.5em}

    \textbf{(c) LLaMA2-13B-Chat}\\[-0.2em]
    \includegraphics[width=0.8\textwidth]{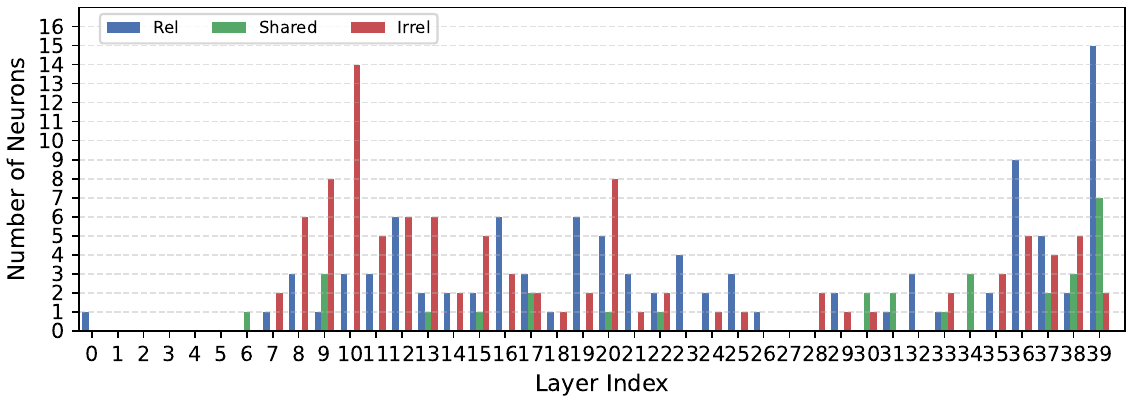}

    \caption{Layer-wise parameter and neuron distributions across models.}
    \label{fig:parameter_distribution}
\end{figure*}

\noindent
Figure~\ref{fig:parameter_distribution} shows the layer-wise distribution of context-aware neurons across 
\textbf{Mistral-7B-Instruct-v0.2}~\cite{jiang2023mistral7b}, 
\textbf{Qwen2.5-14B-Instruct}~\cite{qwen2025qwen25technicalreport}, and 
\textbf{LLaMA-2-13B-Chat}~\cite{touvron2023llama2openfoundation}.
Overall, these model-specific distributions show that context-aware neurons form concentrated layer-wise patterns, with their locations varying across generator models.

\paragraph{Mistral-7B-Instruct-v0.2}
For Mistral-7B-Instruct-v0.2, the identified context-aware neurons show distinct layer-wise patterns.
$\mathcal{P}_{\text{rel}}$ is concentrated around Layers 16--17 and appears again in the final layers, especially around Layers 29--31.
$\mathcal{P}_{\text{irrel}}$ is also prominent around Layers 15--18, with additional occurrences around Layers 20 and 29--31.
$\mathcal{P}_{\text{shared}}$ is sparse in the lower and middle layers but becomes more visible around Layers 29--31.

\paragraph{Qwen2.5-14B-Instruct}
For Qwen2.5-14B-Instruct, context-aware neurons are concentrated predominantly in the upper layers, with few neurons identified in the lower and middle layers.
$\mathcal{P}_{\text{rel}}$ appears mainly after Layer 27, with visible concentrations around Layers 28--31, 36--38, and 40--47.
$\mathcal{P}_{\text{irrel}}$ is also concentrated in the upper layers and forms a pronounced peak around Layers 32--33.
$\mathcal{P}_{\text{shared}}$ is comparatively sparse but appears mainly around Layers 28--31 and 43--47.

\paragraph{LLaMA-2-13B-Chat}
For LLaMA-2-13B-Chat, the identified neurons are distributed across a broader range of layers.
$\mathcal{P}_{\text{irrel}}$ is prominent in the early-to-middle layers, especially around Layers 8--13, and also appears in the upper layers around Layers 36--38.
$\mathcal{P}_{\text{rel}}$ appears intermittently across the middle layers, including Layers 10--12, 16, and 19--22, and becomes highly concentrated in the final layers, particularly around Layers 36 and 39.
$\mathcal{P}_{\text{shared}}$ is less dense overall but becomes more visible around Layers 30--33 and 36--39.

\vspace{0.6em}
\noindent\makebox[\linewidth][c]{%
\begin{minipage}{0.48\textwidth}
    \centering
    \scriptsize
    \setlength{\tabcolsep}{3.5pt}
    \renewcommand{\arraystretch}{1.15}

    \resizebox{\linewidth}{!}{%
    \begin{tabular}{lcccccccc}
        \toprule
        \textbf{Method} &
        \textbf{NQ} &
        \textbf{ASQA} &
        \textbf{TriviaQA} &
        \textbf{SCIQ} &
        \textbf{POPQA} &
        \textbf{HotpotQA} &
        \textbf{2Wiki} &
        \textbf{AVG.} \\
        \midrule
        RALM & 62.21 & 67.82 & 88.60 & 53.30 & 60.85 & 45.44 & 42.27 & 60.07 \\
        RetRobust & 53.23 & 61.92 & 89.80 & 58.20 & 53.24 & 40.84 & 35.17 & 56.06 \\
        InstructRAG & 60.62 & 66.98 & 87.48 & 55.30 & 60.25 & 45.33 & 44.61 & 60.08 \\
        PA-RAG & 67.92 & 74.47 & 89.86 & 56.60 & 64.37 & 49.29 & 46.57 & 64.15 \\
        \cdashline{1-9}
        \textbf{\alg} & & & & & & & & \\
        \hspace{1em}\textit{w/ Contriever} & \underline{69.43} & \underline{74.68} & \underline{91.72} & \underline{60.20} & \underline{66.66} & \underline{51.47} & \underline{46.07} & \underline{65.75} \\
        \hspace{1em}\textit{w/ Splade-v3} & \textbf{69.68} & \textbf{75.31} & \textbf{92.21} & \textbf{60.80} & \textbf{67.37} & \textbf{52.25} & \textbf{46.34} & \textbf{66.28} \\
        \bottomrule
    \end{tabular}}

    \captionof{table}{Results with different retrievers used during neuron mining (SPLADE-v3 vs.\ Contriever), while keeping all other components fixed. Best results are in \textbf{bold} and second-best results are \underline{underlined}.}
    \label{tab:contriever_results}
\end{minipage}%
}
\vspace{0.6em}

\subsection{Sensitivity to Retriever Choice in Neuron Mining.}
\label{appendix:retriever}
To assess the sensitivity of the proposed neuron mining strategy to the choice of retriever, we re-run neuron mining using Contriever~\citep{izacard2021contriever} instead of SPLADE-v3, while keeping all other components fixed. As shown in Table~\ref{tab:contriever_results}, replacing SPLADE-v3 with Contriever yields comparable performance, with \alg{} consistently outperforming all baselines, including \code{PA-RAG}, across benchmarks. These results suggest that the proposed neuron mining strategy is robust to the choice of retriever and scoring scheme.

\begin{table}[h]
\centering
\small
\renewcommand{\arraystretch}{0.8}
\setlength{\tabcolsep}{6pt}
\begin{tabular}{lccc}
\toprule
\textbf{Method} & \textbf{WOW} & \textbf{ELI5} & \textbf{FEVER} \\
\midrule
RALM & 11.80 & 35.63 & 81.30 \\
\textbf{\alg} & \textbf{16.04} & \textbf{43.97} & \textbf{82.41} \\
\bottomrule
\end{tabular}
\caption{Performance comparison on non-QA RAG benchmarks.}
\label{nonqa_results}
\end{table}
\vspace{-4mm}

\subsection{Generalization Beyond QA Benchmarks.}
As shown in Table~\ref{nonqa_results}, our method consistently outperforms \code{RALM} on non-QA RAG benchmarks, including \textsc{WOW}~\cite{dinan2019wizardwikipediaknowledgepoweredconversational}, \textsc{ELI5}~\cite{fan2019eli5longformquestion}, and \textsc{FEVER}~\cite{thorne2018feverlargescaledatasetfact}. 
These results suggest that selective neuron updates enable robust adaptation without degrading general reasoning.

\subsection{Relevant-summary generator robustness.}
We construct the Relevant Summary dataset $\mathcal{D}_{\mathrm{RS}}$ using GPT-4.1-mini. To assess the impact of \alg{} to the choice of relevant-summary generator, we replace GPT-4.1-mini with the open-source Qwen2.5-72B when constructing $\mathcal{D}_{\mathrm{RS}}$. As shown in Table~\ref{tab:rs_generator}, \alg{} consistently outperforms RALM across all benchmarks under this setting, indicating that its effectiveness is independent of a particular model for generating relevant-summary supervision.\begin{table}[H]
    \centering
    \scriptsize
    \setlength{\tabcolsep}{3.8pt}
    \renewcommand{\arraystretch}{1.12}

    \resizebox{\linewidth}{!}{%
    \begin{tabular}{lcccccc}
        \toprule
        \textbf{Method} &
        \textbf{NQ} &
        \textbf{ASQA} &
        \textbf{TriviaQA} &
        \textbf{SCIQ} &
        \textbf{HotpotQA} &
        \textbf{Avg.} \\
        \midrule

        RALM
        & 62.21 & 67.82 & 88.60 & 53.30 & 45.44 & 63.47 \\

        \alg{} w/ Qwen2.5-72B-Instruct
        & \underline{64.23}
        & \underline{69.40}
        & \underline{89.96}
        & \underline{54.31}
        & \underline{46.64}
        & \underline{64.91} \\

        \alg{} w/ GPT-4.1-mini
        & \textbf{68.98}
        & \textbf{71.72}
        & \textbf{92.21}
        & \textbf{60.20}
        & \textbf{48.26}
        & \textbf{68.27} \\

        \bottomrule
    \end{tabular}%
    }

    \caption{
    Effect of the relevant-summary generator.
    Qwen2.5-72B-Instruct is used as an open-source alternative to GPT-4.1-mini while keeping all other components fixed.
    Best results are in \textbf{bold} and second-best results are \underline{underlined}.
    }
    \label{tab:rs_generator}
\end{table}

\begin{table*}[t]
\centering
\small
\setlength{\tabcolsep}{6pt}
\renewcommand{\arraystretch}{1.25}

\arrayrulecolor{black!25}
\begin{tabular}{p{0.14\textwidth} p{0.80\textwidth}}
\arrayrulecolor{black}
\toprule
\multicolumn{2}{c}{\textbf{Examples of Relevant Attribution Set ($\tilde{\mathcal{P}}_{rel}$)}} \\
\midrule
\arrayrulecolor{black!25}

\textit{c} &
Context: Arthur's Lady's Home Magazine. Arthur's Lady's Home Magazine Arthur's Home Magazine (1852--ca.1898) or Ladies' Home Magazine was an American periodical published in Philadelphia by Timothy Shay Arthur. Editors Arthur and Virginia Francis Townsend selected writing and illustrations intended to appeal to female readers. Among the contributors: Mary Tyler Peabody Mann and Kate Sutherland. In its early years the monthly comprised a selection of articles originally published in Arthur's weekly ``Home Gazette.'' Its nonfiction stories contained occasional factual inaccuracies for the sake of a good read. →

Question: Which magazine was started first Arthur's Magazine or First for Women?

Proposed Answer: Arthur's Magazine \\
\hline
\textit{q} &
If the proposed answer can be derived by referring to the context, answer YES; otherwise, answer NO.
The correct answer is \\
\hline
choices & ['YES', 'NO'] \\
\hline
gold answer & 0 \\

\arrayrulecolor{black}
\midrule
\arrayrulecolor{black!25}

\textit{c} &
Context: The Oberoi Group. The Oberoi Group is a hotel group with its head office in Delhi. Founded in 1934, the company owns and/or operates 35 luxury hotels and two river cruise ships in six countries, primarily under its Oberoi Hotels \& Resorts and Trident Hotels brands. The foundations of the Oberoi Group date back to 1934 when Rai Bahadur Mohan Singh Oberoi bought two properties in Delhi and Shimla. →

Question: The Oberoi family is part of a hotel company that has a head office in what city?

Proposed Answer: Delhi \\
\hline
\textit{q} &
If the proposed answer can be derived by referring to the context, answer YES; otherwise, answer NO.
The correct answer is \\
\hline
choices & ['YES', 'NO'] \\
\hline
gold answer & 0 \\

\arrayrulecolor{black}
\midrule
\multicolumn{2}{c}{\textbf{Examples of Irrelevant Attribution Set ($\tilde{\mathcal{P}}_{irrel}$)}} \\
\midrule
\arrayrulecolor{black!25}

\textit{c} &
Context: Autism. Sustained special education programs and behavior therapy early in life can help children acquire self-care, communication, and job skills, and often improve functioning and decrease symptom severity and maladaptive behaviors. Educational interventions often used include applied behavior analysis (ABA), speech and language therapy, social skills therapy, and occupational therapy. →

Question: Which magazine was started first Arthur's Magazine or First for Women?

Proposed Answer: Arthur's Magazine \\
\hline
\textit{q} &
If the proposed answer can be derived by referring to the context, answer YES; otherwise, answer NO.
The correct answer is \\
\hline
choices & ['YES', 'NO'] \\
\hline
gold answer & 0 \\

\arrayrulecolor{black}
\midrule
\arrayrulecolor{black!25}

\textit{c} &
Context: Albedo. Albedo is the measure of the diffuse reflection of solar radiation out of the total solar radiation received by an astronomical body. Enceladus has one of the highest known albedos in the Solar System, reflecting about 99\% of incident radiation. →

Question: The Oberoi family is part of a hotel company that has a head office in what city?

Proposed Answer: Delhi \\
\hline
\textit{q} &
If the proposed answer can be derived by referring to the context, answer YES; otherwise, answer NO.
The correct answer is \\
\hline
choices & ['YES', 'NO'] \\ 
\hline
gold answer & 0 \\

\arrayrulecolor{black}
\bottomrule
\end{tabular}

\caption{Case \uppercase\expandafter{\romannumeral1}: Representative examples of the binary decision dataset used for Integrated Gradients-based neuron attribution. The dataset frames the task as a forced binary choice (YES/NO) to determine if the answer is strictly derivable from the context. This allows us to disentangle neurons responsible for processing relevant contexts ($\tilde{\mathcal{P}}_{rel}$) from those processing irrelevant ones ($\tilde{\mathcal{P}}_{irrel}$).}
\label{tb:ig_dataset_examples}
\end{table*}

\begin{table*}[h!]
\centering
\resizebox{1.0\textwidth}{!}{
\begin{tabular}{b|p{0.95\textwidth}}
\thickhline

\textbf{Question} &
Which magazine was started first Arthur's Magazine or First for Women? \\ \hline

\textbf{Document 1} &
Autism. sustained special education programs and behavior therapy early in life can help children acquire self-care, communication, and job skills, and often improve functioning and decrease symptom severity and maladaptive behaviors; claims that intervention by around age three years is crucial are not substantiated. While medications have not been found to help with core symptoms, they may be used for associated symptoms, such as irritability, inattention, or repetitive behavior patterns. Section:Management.:Education. Educational interventions often used include applied behavior analysis (ABA), developmental models, structured teaching, speech and language therapy, social skills therapy, and occupational therapy. Among these approaches, interventions either treat autistic \\ \hline

\textbf{Document 2} &
Albedo. back into space than what they absorb, effectively cooling the Earth. This has been a concern since arctic ice and snow has been melting at higher rates due to higher temperatures, creating regions in the arctic that are notably darker (being water or ground which is darker color) and reflects less heat back into space. This feedback loop results in a reduced albedo effect. Section:Examples of terrestrial albedo effects.:Climate and weather. Albedo affects climate by determining how much radiation a planet absorbs. The uneven heating of Earth from albedo variations between land, ice, or ocean surfaces can drive weather. Section:Examples \\ \hline

\textbf{Document 3} &
Autism. social-communication skills in young children, although there is less evidence in its treatment of global symptoms. Neuropsychological reports are often poorly communicated to educators, resulting in a gap between what a report recommends and what education is provided. It is not known whether treatment programs for children lead to significant improvements after the children grow up, and the limited research on the effectiveness of adult residential programs shows mixed results. The appropriateness of including children with varying severity of autism spectrum disorders in the general education population is a subject of current debate among educators and researchers. Section:Management.:Medication. Medications may \\ \hline

\textbf{Document 4} &
Albedo. the mean temperature of the planet would drop to about 0 °C. In contrast, if the entire Earth was covered by water — a so-called ocean planet — the average temperature on the planet would rise to almost 27 °C. Section:Terrestrial albedo.:White-sky, black-sky, and blue-sky albedo. For land surfaces, it has been shown that the albedo at a particular solar zenith angle ``$\theta$'' can be approximated by the proportionate sum of two terms: with formula\_3 being the proportion of direct radiation from a given solar angle, and formula\_4 being the proportion of diffuse illumination, the actual albedo formula\_5 (also called \\ \hline

\textbf{Document 5} &
Autism. and the educational system are the main resources for treatment. Services should be carried out by behavior analysts, special education teachers, speech pathologists, and licensed psychologists. Studies of interventions have methodological problems that prevent definitive conclusions about efficacy. However, the development of evidence-based interventions has advanced in recent years. Although many psychosocial interventions have some positive evidence, suggesting that some form of treatment is preferable to no treatment, the methodological quality of systematic reviews of these studies has generally been poor, their clinical results are mostly tentative, and there is little evidence for the relative effectiveness of treatment options. Intensive, \\ \hline


\textbf{Relevant summary} &
EOT(End of Text) \\ \thickhline

\end{tabular}
}
\caption{Case \uppercase\expandafter{\romannumeral2}: An example of the Irrelevant Context Denoising Dataset. When provided with retrieved documents that are entirely irrelevant to the query, the model is trained to output an End-of-Text (EOT) token, enforcing the suppression of noise.}
\label{tb:rs_example1}
\end{table*}

\begin{table*}[h!]
\centering
\resizebox{1.0\textwidth}{!}{
\begin{tabular}{b|p{0.95\textwidth}}
\thickhline

\textbf{Question} &
Which magazine was started first Arthur's Magazine or First for Women? \\ \hline

\textbf{Document 1} &
Arthur's Lady's Home Magazine. Arthur's Lady's Home Magazine Arthur's Home Magazine (1852-ca.1898) or Ladies' Home Magazine was an American periodical published in Philadelphia by Timothy Shay Arthur. Editors Arthur and Virginia Francis Townsend selected writing and illustrations intended to appeal to female readers. Among the contributors: Mary Tyler Peabody Mann and Kate Sutherland. In its early years the monthly comprised a selection of articles originally published in Arthur's weekly ``Home Gazette.'' Its nonfiction stories contained occasional factual inaccuracies for the sake of a good read. A contemporary review judged it ``gotten up in good taste and well; and is in nothing overdone. Even its \\ \hline

\textbf{Document 2} &
Arthur's Magazine. Arthur's Magazine Arthur's Magazine (1844--1846) was an American literary periodical published in Philadelphia in the 19th century. Edited by Timothy Shay Arthur, it featured work by Edgar A. Poe, J.H. Ingraham, Sarah Josepha Hale, Thomas G. Spear, and others. In May 1846 it was merged into ``Godey's Lady's Book''. A few years later Arthur would launch a new publication entitled ``Arthur's Home Magazine''. Section:Further reading. Section:External links. \\ \hline

\textbf{Document 3} &
Arthur (magazine). Arthur (magazine) Arthur magazine was a bi-monthly periodical that was founded in October 2002, by publisher Laris Kreslins and editor Jay Babcock. It received favorable attention from other periodicals such as ``L.A. Weekly'', ``Print'', ``Punk Planet'' and ``Rolling Stone''. ``Arthur'' featured photography and artwork from Spike Jonze, Art Spiegelman, Susannah Breslin, Gary Panter and Godspeed You! Black Emperor. Arthur's regular columnists included Byron Coley, Thurston Moore, Daniel Pinchbeck, Paul Cullum, Douglas Rushkoff, and T-Model Ford. ``Arthur'' magazine was particularly drawn to noise music, stoner metal, folk and other types of psychedelia. The first issue of ``Arthur'' featured an interview with \\ \hline

\textbf{Document 4} &
Marianne Ehrmann. stopped the further issues. Section:Life and career.:Amaliens Erholungsstunden. ``Amaliens Erholungsstunden'' was the first women's magazine, and it was self-published monthly by Marianne Ehrmann under her own name. She sent advance notices to acquaintances, friends and potential patrons to distribute these in the circle, and the first six editions resulted in an increase of subscribers. The debts contracted by the self-publishing were indeed paid off, but in all a financial disaster, that's why in January 1791 Marianne and her husband started a cooperation with the publishing house ``J. G. Cottaische Verlagsbuchhandlung'' in Tübingen. Marianne Ehrmann thought to concentrate on the journalistic \\ \hline

\textbf{Document 5} &
Arthur Jewitt. ``The Northern Star, or Yorkshire Magazine,'' a monthly register of arts, biography, statistics, manufactures, \&c., which ran to three volumes, 1817--18. On 1 January 1818 he brought out the first number of ``The Sylph, or Lady's Magazine for Yorkshire, Derbyshire, and the adjoining Counties.'' ``The Lincoln and Lincolnshire Cabinet and Annual Intelligencer'' appeared at Lincoln during 1827--9. His ``Matlock Companion,'' 1835, and ``Derbyshire Gems'' were very popular. His ``Handbook of Practical Perspective,'' 1840, and his ``Handbook of Geometry,'' 1842, were adopted by the committee of council on education. He contributed mathematical papers to the ``British Diary'' and to the ``Lady's'' \\ \hline

\textbf{Relevant summary} &
The document mentions ``Arthur's Magazine'' as an American literary periodical published from 1844 to 1846. It was edited by Timothy Shay Arthur and later merged into ``Godey's Lady's Book'' in 1846. It also references ``Arthur's Home Magazine'' starting in 1852, but no start date or details are given for ``First for Women.'' There is no mention of ``First for Women'' magazine's start date or any comparison to Arthur's Magazine. \\ \bottomrule

\end{tabular}
}
\caption{Case \uppercase\expandafter{\romannumeral3}: An example of the Relevant Summary Dataset ($\mathcal{D}_{RS}$). The external LLM synthesizes a concise summary using only the query-aligned documents, while effectively filtering out unrelated information.}
\label{tb:rs_example2}
\end{table*}

\end{document}